\documentclass[runningheads]{llncs}
\newcommand{\ARXIV}[2]{#1} 
 
\PassOptionsToPackage{table,dvipsnames}{xcolor}

\usepackage{eccv}

\usepackage{eccvabbrv}

\usepackage{graphicx}
\usepackage{booktabs}
\usepackage{scalerel}

\usepackage[accsupp]{axessibility}  

\usepackage[pagebackref,breaklinks,colorlinks,citecolor=eccvblue]{hyperref}

\usepackage{orcidlink}

\usepackage{graphicx}
\usepackage{amsmath}
\usepackage{amssymb}
\usepackage{times}
\usepackage{mathtools}
\usepackage{mathrsfs}  
\usepackage{amssymb}
\usepackage{booktabs}
\usepackage{multirow,multicol}
\usepackage{array}
\usepackage{tabularx}
\usepackage{balance}
\usepackage{placeins}
\usepackage{subcaption}

\usepackage{fontawesome5}
\usepackage{pifont}
\usepackage{enumitem}
\usepackage{tikz}
\usepackage{pgfplots}
\usepackage{styles}
\usepackage{caption}
\usepackage[outline]{contour}
\usepackage{microtype}
\usetikzlibrary{tikzmark,patterns,shapes.misc,shapes.geometric,positioning,fit}

\usepackage[hang,flushmargin]{footmisc}
\usepackage[accsupp]{axessibility}  

\setlist[itemize]{noitemsep, topsep=1pt}
\usepackage[font=small,skip=4pt]{caption}

\begin{document}

\title{
OmniSat: 
Self-Supervised Modality Fusion\\ for Earth Observation
}

\titlerunning{Cross-Modal Token Alignment}

\author{
    Guillaume Astruc \textsuperscript{1,3,4}\orcidlink{0009-0008-6242-0939}
    \and
    Nicolas Gonthier \textsuperscript{1,2}\orcidlink{0009-0008-6242-0939}
    \and\\
    Clement Mallet \textsuperscript{1}\orcidlink{0009-0008-6242-0939}
    \and
    Loic Landrieu\textsuperscript{1,3}\orcidlink{0009-0008-6242-0939}
    }
\institute{
    {\textsuperscript{1} Univ Gustave Eiffel, IGN, ENSG, LASTIG, France} \qquad
     {\textsuperscript{2} IGN, France}\\
    {\textsuperscript{3} LIGM, Ecole des Ponts, Univ Gustave Eiffel, CNRS, France}
     \qquad
     {\textsuperscript{4} CNES, France}
}
\authorrunning{G.~Astruc et al.}

\maketitle

\begin{abstract}
The diversity and complementarity of sensors available for Earth Observations (EO) calls for developing bespoke self-supervised multimodal learning approaches.
However, current multimodal EO datasets and models typically focus on a single data type, either mono-date images or time series, which limits their impact. To address this issue, we introduce OmniSat, a novel architecture able to merge diverse EO modalities into expressive features without labels by exploiting their alignment.
To demonstrate the advantages of our approach, we create two new multimodal datasets by augmenting {existing ones} with new modalities. As demonstrated for three downstream tasks---forestry, land cover classification, and crop mapping---OmniSat can learn rich representations without supervision, leading to state-of-the-art performances in semi- and fully supervised settings. Furthermore, our multimodal pretraining scheme improves performance even when only one modality is available for inference. The code and dataset are available at \GITHUB.
\end{abstract}
\setlength{\parskip}{-0.13em}

\section{Introduction}

Self-supervised multimodal learning has recently gathered significant interest within computer vision \cite{zong2023self,girdhar2023imagebind,shukor2023unival}. Earth Observation (EO) is particularly well-suited for developing and evaluating such approaches \cite{fuller2023croma,irvin2023usat}, thanks to the large amount of open-access data captured by sensing technologies with complementary capabilities \cite{ghamisi2019multisource,schmitt2016data}. 
Combining different sources of EO observations is crucial for several high-impact applications, including environmental \cite{coppin2002digital,secades2014earth,skidmore2021priority} and climate monitoring \cite{yang2013role,lacoste2021toward}, as well as improving food security \cite{nakalembe2020urgent}. Moreover, learning with few or no labels is essential for developing regions with limited data annotation capabilities \cite{kuffer2020role,anderson2017earth,mai2023opportunities}.

Despite this potential, most multimodal EO datasets and models focus on a single data type, either mono-date images or time series. This limitation prevents them from simultaneously leveraging the spatial resolution of aerial images \cite{li2012current,manfreda2018use}, the temporal and spectral resolutions of optical satellite time series \cite{Sentinel-2-paper}, and the resilience of radar to weather effects \cite{moreira2013tutorial,amitrano2021earth}. 
Additionally, existing approaches are often limited for a given set of sensors, limiting their applicability.

To address these challenges, we introduce OmniSat, a novel architecture designed for the self-supervised fusion of diverse EO data. 
Existing multimodal approaches often map multiple unrelated observations from different modalities to one pivot modality \cite{shukor2023unival,girdhar2023imagebind} or a shared latent space \cite{girdhar2022omnivore,srivastava2024omnivec}. In contrast, OmniSat merges multiple views of the same area from different modalities into a single representation combining the specific information of each modality \cite{recasens2023zorro,greenwell2024watch,benedetti2018m}.

In computer vision, obtaining finely aligned multimodal observations generally requires specialized sensors \cite{liao2022kitti,Silberman:ECCV12,krispel2020fuseseg} or the computation of complex mappings between modalities \cite{robert2022learning,dai20183dmv}. 
On the other hand, EO data can be naturally aligned with georeferencing. 
To leverage this property, we adapt multimodal contrastive learning \cite{radford2021learning,huang2022mavil} and cross-modal masked auto-encoding techniques \cite{hackstein2024exploring} to learn rich multimodal EO representations with a generalist fusion scheme and without annotations. 

To address the scarcity of EO datasets with a diverse range of heterogeneous modalities (see \cref{tab:datasets}), we enrich the TreeSatAI  \cite{ahlswede2022treesatai} and PASTIS-R \cite{garnot2021panoptic,garnot2022multi} datasets with new aligned modalities. This allows us to evaluate OmniSat's ability to handle an arbitrary number of inputs with varying natures and resolutions.
Our contributions can be summarized as follows:
\begin{itemize}
    \item We introduce OmniSat, a new model that learns to combine varied sources of EO observations in a self-supervised manner, resulting in richer joint representations that capture the unique characteristics of each modality.
    \item We augment {two EO benchmarks to create the first datasets} 
    with three modalities of different natures (very high resolution images, optical and SAR time series).
    \item We demonstrate that OmniSat can leverage diverse modalities to learn rich representations, establishing new states-of-the-art for tree species, crop type, and land cover classification. Furthermore, our cross modal self-supervised training scheme improves performance even when only one modality is available during inference.
\end{itemize}

\begin{table}[t]
\caption{\textbf{Publicly Available Multimodal EO Datasets.} We provide in parenthesis the spatial resolutions of the single-date images and labels, and the temporal resolutions of time series. S1/S2 denotes Sentinel-1 and 2. \raisebox{0.3mm}{\small $\star$}\small~: \textbf{modalities added in this work}.\vspace{-1mm}}
\label{tab:datasets}
\centering
\scriptsize
\renewcommand{\arraystretch}{1.5}
\setlength{\tabcolsep}{2pt}
\begin{tabular}{llll}
\toprule
\multirow{2}{*}{Dataset} & \multicolumn{2}{c}{Modalities}  &  \multirow{2}{*}{\;\;Labels}\\
     \cline{2-3}
& images (single date)& time series & \\
\midrule
SpaceNet6 \cite{shermeyer2020spacenet} & SAR+optical (0.5m-2m) & \qquad\xmark  & building footprint (<1m)\\
TreeSatAI \cite{ahlswede2022treesatai} & aerial + S1/S2 (0.2-10m) & \qquad\xmark & forestry (60m) \\
BigEarthNet \cite{sumbul2021bigearthnet} & S1/S2 (10m) & \qquad\xmark  & land cover (100m) \\
DFC20 \cite{robinson2021global} & S1/S2 (10m) &  \qquad\xmark  & land cover (500m)\\
MDAS \cite{hu2022mdas} & S1/S2 + hyperspectral (2.2-10m) &  \qquad\xmark & land cover (0.25m) \\
DOFA \cite{xiong2024dofa} & NAIP + Gaofen + S1/S2 + EnMAP (1-30m)& \qquad\xmark & \qquad\xmark \\\greyrule
PASTIS-R \cite{garnot2021panoptic,garnot2022multi} & 
\qquad\xmark & S1/S2 (30-140 / year) & agriculture (10m) \\
SSL4EO-S12 \cite{wang2022ssl4eo} & \qquad\xmark & S1/S2 (4 / year) & \qquad\xmark\\
DFC21-DSE \cite{ma2021outcome} & \qquad\xmark & S1/S2 + LS8 (3-9/year) & 
human activity (500m)\\
MapInWild \cite{ekim2023mapinwild} & \qquad\xmark & S1/S2 (4 / years) & protected areas (10m)\\
SEN12MS-CR-TS \cite{ebel2022sen12ms} & \qquad\xmark & S1/S2 (30 / years) & cloud cover (10m) \\
MultiSenGE \cite{wenger2022multisenge} &  \qquad\xmark & S1/S2 (30-140 / years) & land cover (10m) 
\\\greyrule
FLAIR \cite{garioud2023flair} & aerial (0.2m) & S2 (20-114 / year) & land cover (0.2m) \\
Satlas \cite{bastani2023satlaspretrain} & NAIP (0.5 -2m) & S2 (8-12 / year) & various \\
\textbf{PASTIS-HD} & \bf \raisebox{0.3mm}{\small $\star$} SPOT 6-7 (1.5m) & S1/S2 (30-140 / year) & agriculture (10m) \\
\textbf{TreeSatAI-TS} & aerial (0.2m) & \bf \raisebox{0.3mm}{\small $\star$} S1/S2 (10-70 / year)  & forestry (60m)
\\\bottomrule
\end{tabular}
\end{table}


\section{Related Work}

This section provides an overview of self-supervised and multimodal learning, emphasizing the specificities of their usage for Earth observation. Lastly, we highlight the scarcity of multimodal EO datasets with diverse data types.

\parag{\bf Self-Supervised Learning.}
This technique consists in learning expressive data representations without labels by using a pretext task. This approach has been particularly successful for natural language \cite{kenton2019bert} and image \cite{oquab2023dinov2} analysis.  
Initially focused on discriminative tasks \cite{gidaris2018unsupervised,noroozi2016unsupervised,zhang2016colorful}, recent self-supervised approaches for images can be categorized as contrastive or generative.

\emph{Contrastive methods} minimize the distance between representations of paired samples, often the same image under different transformations, and maximize the distance with other samples \cite{chen2020simple,he2020momentum,caron2020unsupervised}.
More efficient methods only consider positive samples and avoid mode collapse by introducing various asymmetries
\cite{grill2020bootstrap,chen2021exploring} or normalization \cite{caron2021emerging}.
Such approaches have been successfully adapted to EO, for which samples are paired according to their location \cite{tseng2022croco} or time of acquisition \cite{ayush2021geography,manas2021seasonal}.

\begin{figure}[t]
    \centering
     \centering \scriptsize{
    \begin{tabular}{ccc}
    VHR aerial 0.2 m & VHR aerial 0.2 m & \large{$\star$}\scriptsize \textbf{VHR satellite 1.5 m} \\
\includegraphics[width=.2\linewidth]{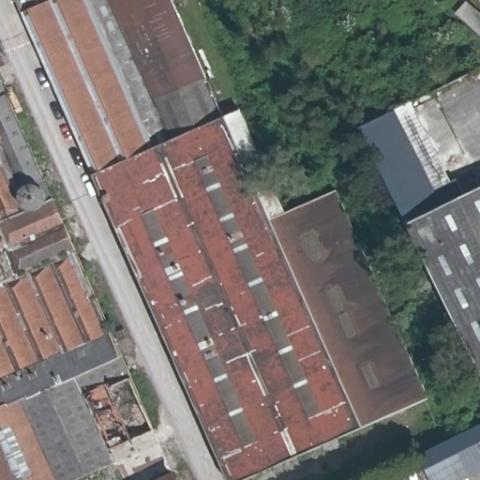}
&
\includegraphics[angle=180,width=.2\linewidth,origin=c]{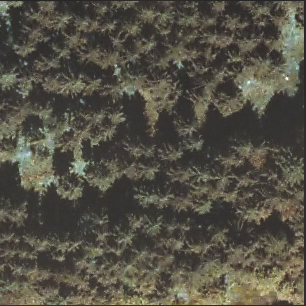}
&
\includegraphics[width=.2\linewidth]{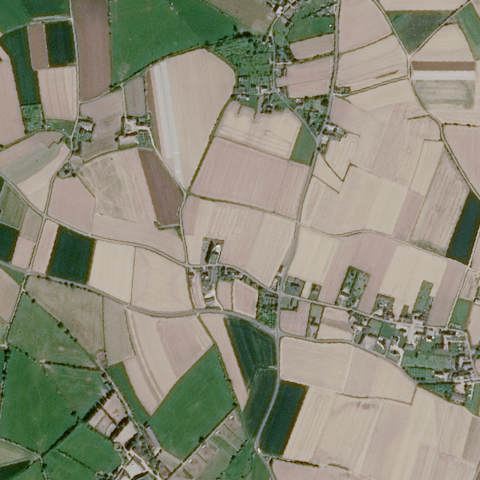}
\\
Sentinel-2 time series
&
\large{$\star$}\scriptsize~Sentinel-2 \bf{time series}
&
PASTIS: Sentinel-2 time series
\\
\multirow{3}{*}[2mm]{ 
\begin{tabular}{cc}
\includegraphics[width=.07\linewidth]{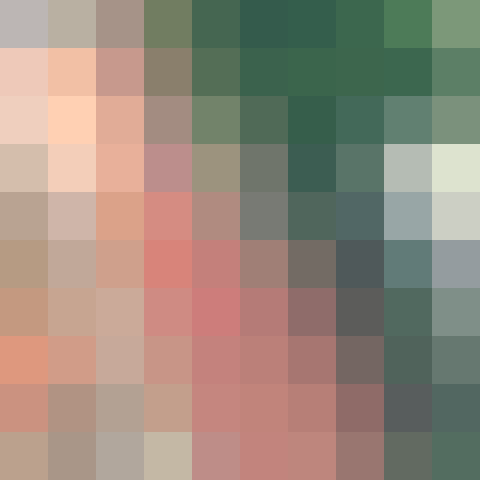}&
\includegraphics[width=.07\linewidth]{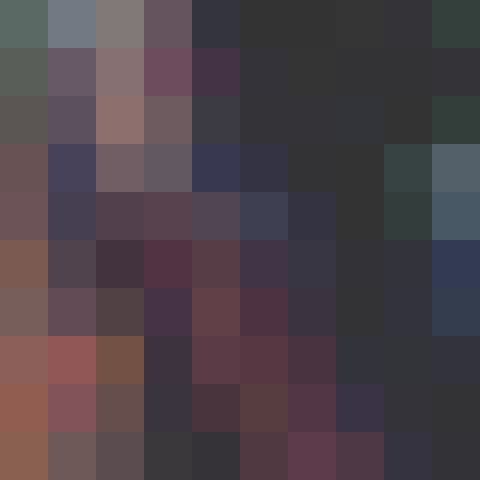}\\           \includegraphics[width=.07\linewidth]{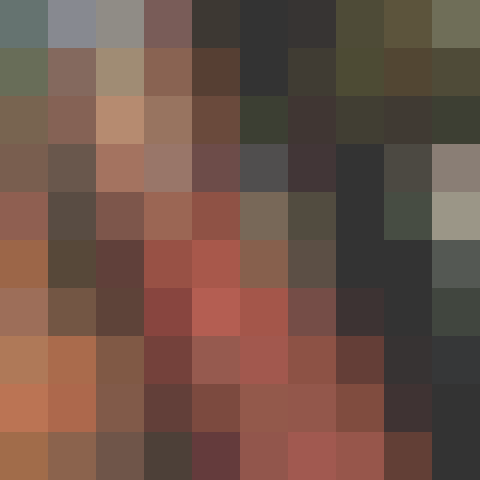}&
\includegraphics[width=.07\linewidth]{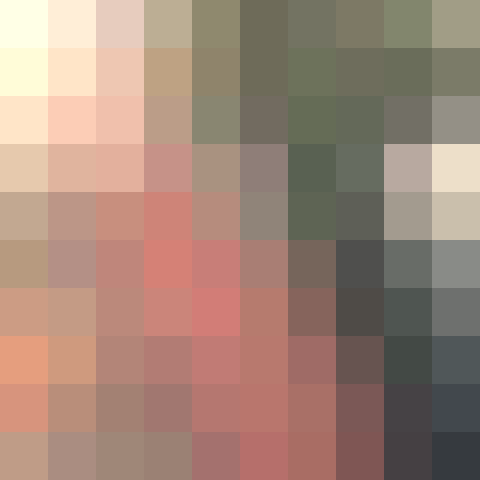}
\end{tabular}
}
&
\begin{tabular}{m{.06\linewidth}@{\,}m{.03\linewidth}@{\,}m{.06\linewidth}@{\,}m{.06\linewidth}@{\,}m{.06\linewidth}}
    \includegraphics[width=1\linewidth, height=.038\textheight]{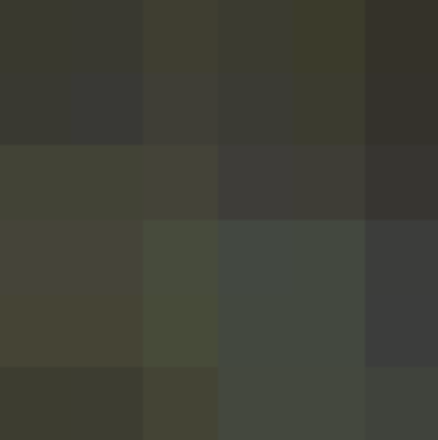}
        &
        \contour{black}{$\rightarrow$}
        &
       \includegraphics[width=1\linewidth, height=.038\textheight]{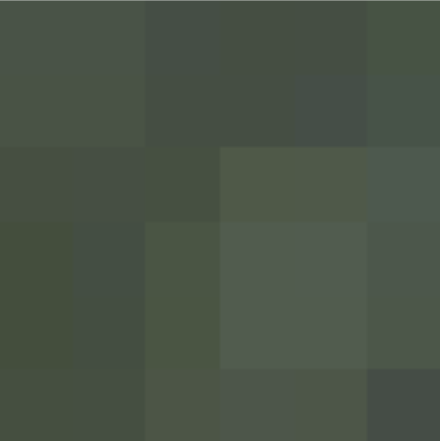}
       &
       \includegraphics[width=1\linewidth, height=.038\textheight]{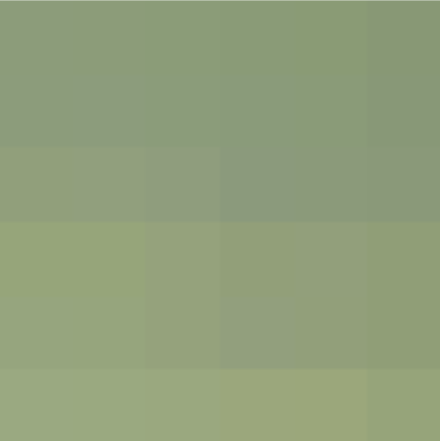}
       &
       \includegraphics[width=1\linewidth, height=.038\textheight]{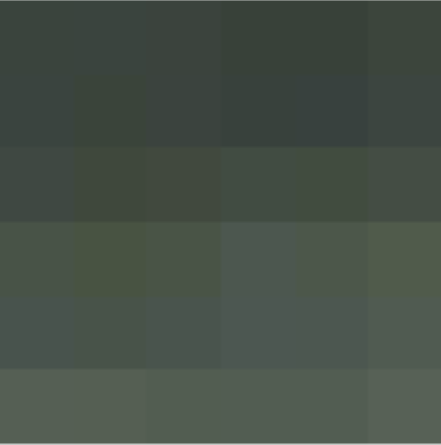}
\end{tabular}
&
 \begin{tabular}{c@{\,}c@{\,}c@{\,}c}
        \includegraphics[width=.06\linewidth, height=.038\textheight]{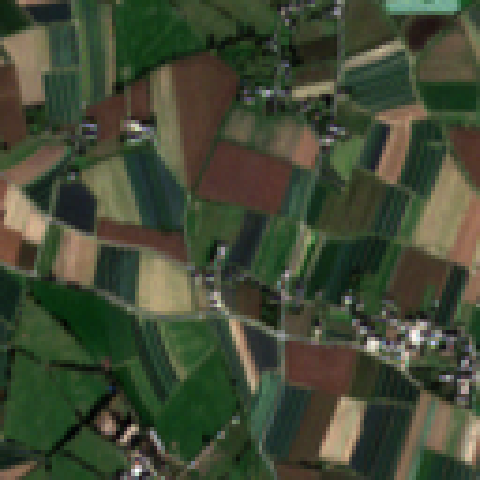}&        \includegraphics[width=.06\linewidth, height=.038\textheight]{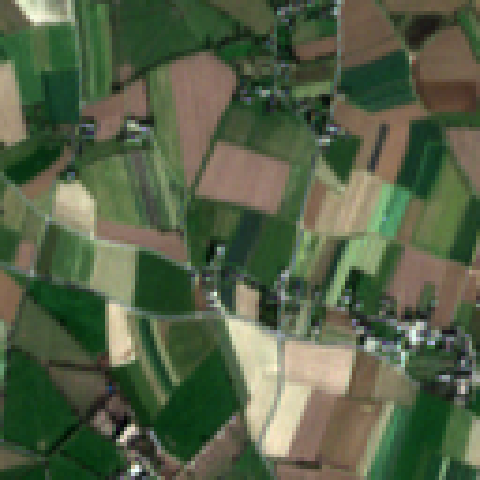} &       \includegraphics[width=.06\linewidth, height=.038\textheight]{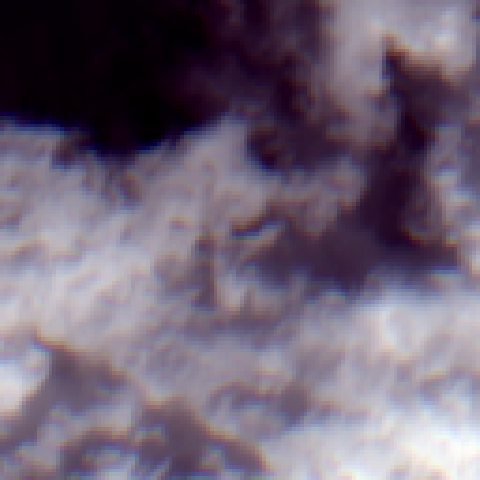} &       \includegraphics[width=.06\linewidth, height=.038\textheight]{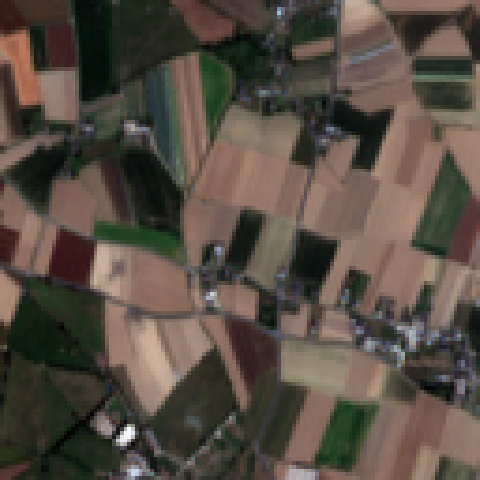}
        \end{tabular}
\\
&
{\large{$\star$}\scriptsize}~Sentinel-1 \bf{time series}
&
\scriptsize{PASTIS-R: Sentinel-1 time series}
\\
&\begin{tabular}{m{.06\linewidth}@{\,}m{.03\linewidth}@{\,}m{.06\linewidth}@{\,}m{.06\linewidth}@{\,}m{.06\linewidth}}
        \includegraphics[width=1\linewidth, height=.038\textheight]{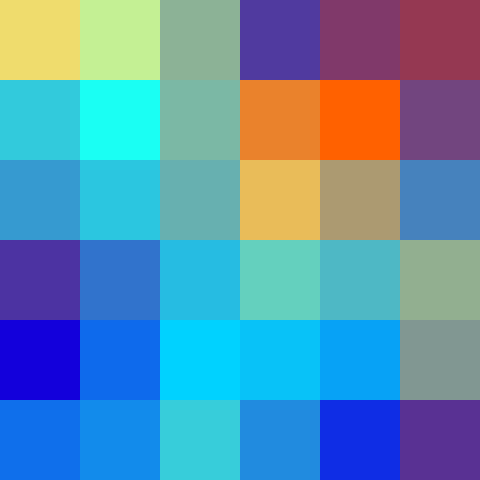}
        &
        \contour{black}{$\rightarrow$}
        &
       \includegraphics[width=1\linewidth, height=.038\textheight]{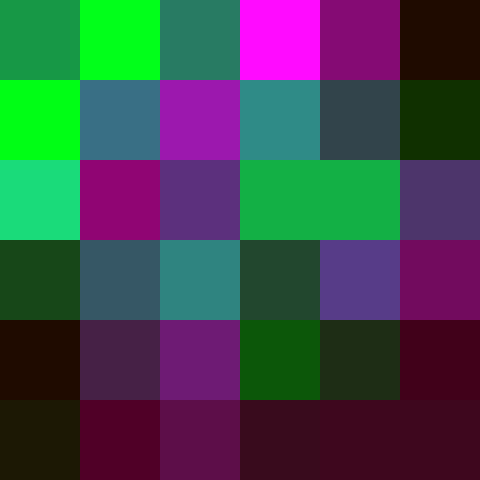}
       &
       \includegraphics[width=1\linewidth, height=.038\textheight]{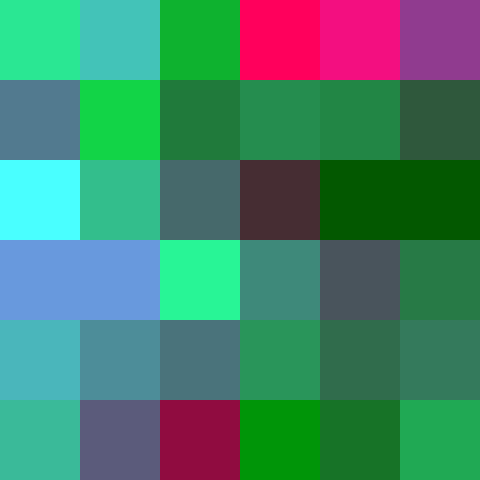}
       &
       \includegraphics[width=1\linewidth, height=.038\textheight]{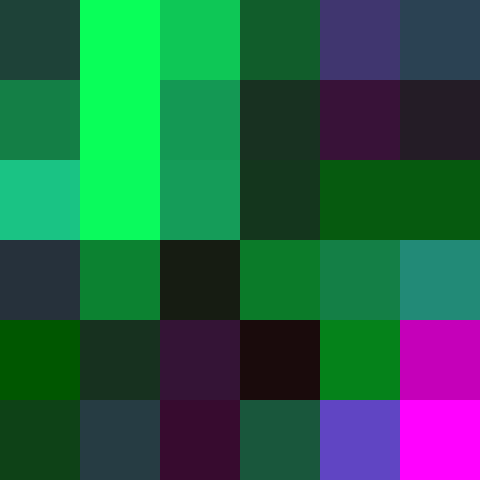}
\end{tabular}
&
         \begin{tabular}{c@{\,}c@{\,}c@{\,}c}   
    \includegraphics[width=.06\linewidth, height=.038\textheight]{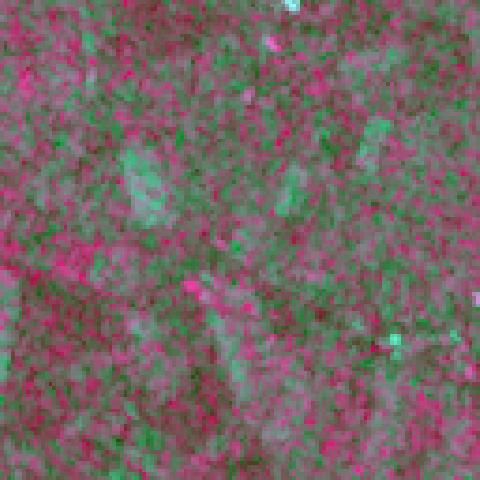}
        &
    \includegraphics[width=.06\linewidth, height=.038\textheight]{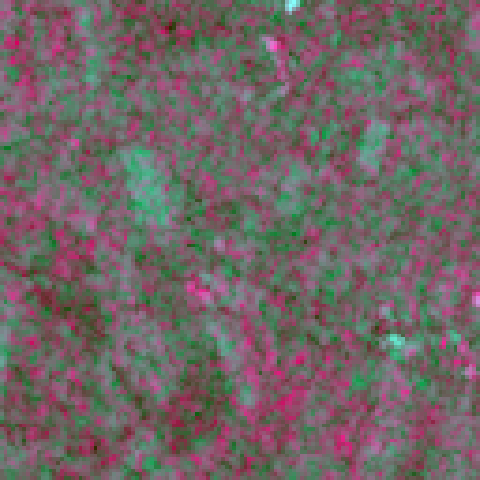}
        &
    \includegraphics[width=.06\linewidth, height=.038\textheight]{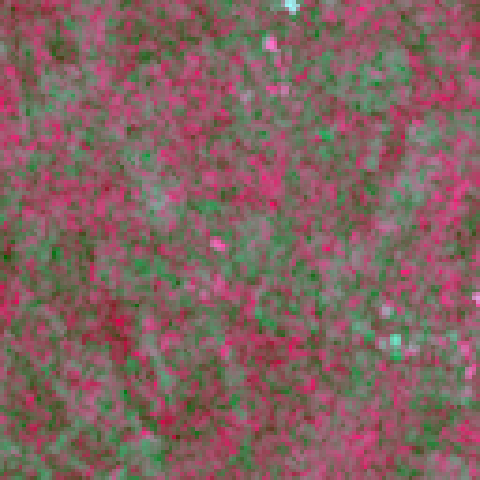}
        &
    \includegraphics[width=.06\linewidth, height=.038\textheight]{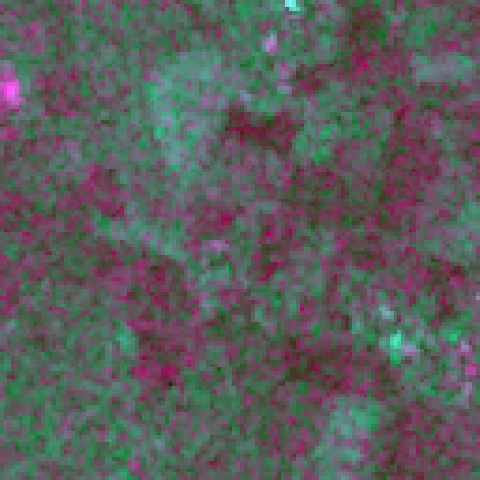}
    \end{tabular}
\\
\begin{subfigure}{0.3\textwidth}
 \caption{FLAIR}
\label{fig:data:flair}
\end{subfigure}
&
\begin{subfigure}{0.3\textwidth}
\caption{\bf TreeSatAI-TS}
\label{fig:data:treesat}
\end{subfigure}
&
\begin{subfigure}{0.3\textwidth}
\caption{\bf PASTIS-HD}
\label{fig:data:pastis}
\end{subfigure}
    \end{tabular}
    }
    \caption{{\bf Datasets.} We represent {three} tiles from the considered multilabel classification datasets:
    FLAIR (\subref{fig:data:flair}), TreeSatAI-TS (\subref{fig:data:treesat}) {and PASTIS-HD (\subref{fig:data:pastis}}). TreeSatAI-TS is a new dataset built by replacing the single-date Sentinel-1 and 2 images of TreeSatAI \cite{ahlswede2022treesatai} by year-long time series. {PASTIS-HD (\subref{fig:data:pastis}) adds VHR satellite images to PASTIS-R \cite{garnot2022multi}}. 
    $\star$~: {\bf modalities added in this work}.
     }
    \label{fig:data}
\end{figure}

\emph{Generative methods} reason at the level of individual token---a small portion of the input, typically a patch for images  \cite{dosovitskiy2020image}. The objective is to reconstruct the masked tokens of an input image in pixel \cite{he2022masked,bao2021beit,xie2022simmim} or feature space \cite{assran2023self}.
This principle has been successfully adapted to EO analysis \cite{gao2022general,cong2022satmae,yuan2022sits}, and was further extended to handle multiple spatial scales \cite{reed2023scale},  multimodality \cite{fuller2023croma,irvin2023usat}, or hyperspectral observations \cite{ibanez2022masked,liu2022band}.

Several hybrid approaches combine the discriminative power of contrastive methods and the scalability of generative objectives for natural images \cite{oquab2023dinov2,zhou2022image} and EO data \cite{fuller2023croma}. 
Our proposed OmniSat model also implements both mechanisms. A key feature is that we leverage the precise alignment between different sources of EO data to contrastively match small patches of different modalities rather than entire images or time series.

\parag{\bf Self-Supervised Multimodal Learning.}
Multimodal computer vision has received a lot of interest \cite{bayoudh2022survey}, notably due to the success of cross-modal pre-training \cite{radford2021learning}. Recent models align the embeddings of heterogeneous modalities such as video and sound \cite{huang2022mavil}, depth and images \cite{hazirbas2017fusenet}, text and image \cite{baevski2022data2vec,alayrac2022flamingo}, or multiple combinations of these modalities \cite{shukor2023unival,girdhar2023imagebind,girdhar2022omnivore,srivastava2024omnivec}.

Multimodal learning also has a long history in EO \cite{yang2021muti,li2022deep,pohl1998review} due to the large variety and complementarity of sensors \cite{ghamisi2019multisource,schmitt2016data}.
However, recent transformer-based architectures \cite{vaswani2017attention} for EO are often limited to one type of modality, be it a single image \cite{cong2022satmae,reed2023scale} or time-series \cite{tarasiou2023vits,garnot2022multi}. For example, CROMA \cite{fuller2023croma} and PRESTO \cite{tseng2023lightweight}  are specifically designed for paired optical and radar observations, but cannot handle Very High Resolution (VHR) data. USat \cite{irvin2023usat} considers images with different resolutions, but only takes a single date within a time series.
UT\&T \cite{garioud2023flair} can natively take single and multi-date observations of different modalities, but cannot be easily pre-trained in a self-supervised manner since it relies on convolutions and an ad-hoc late fusion scheme.

\parag{\bf Multimodal EO Datasets.}
As reported in \tabref{tab:datasets}, many multimodal EO datasets use Sentinel-1 \cite{BAO202386} and 2 \cite{Sentinel-2-paper} data for applications ranging from land cover to forestry analysis and fire detection. We also note that most multimodal datasets only contain data of one type: mono-date image or time series.
Several datasets (BigEarthNet \cite{sumbul2021bigearthnet}, DFC20 \cite{robinson2021global}, MDAS \cite{hu2022mdas}) select a single date from time series. However, single Sentinel-1 and 2 acquisitions can be significantly affected by rain and cloud cover, respectively. Furthermore, capturing the temporal dynamics is crucial to characterize the phenology of vegetation \cite{vrieling2018vegetation},

FLAIR \cite{garioud2023flair} is the first multimodal EO dataset to propose both very high spatial resolution ($\leq 2$m) and high temporal resolution ($>4$ images/year).
Satlas \cite{bastani2023satlaspretrain} combines Sentinel-2 time series and for 5\% to tiles (continental US), very high definition NAIP images.
The functional map of the World \cite{christie2018functional} integrates observations from various sensors, but most areas are only observed with one sensor.
Two other datasets contain time series and single images from multiple sources, but were not available at the time of writing this article:
IARPA-SMART \cite{goldberg2023automated} and DOFA \cite{xiong2024dofa}.

To showcase how OmniSat can consume an arbitrary number of modalities with different spatial, spectral, and temporal resolutions, we selected two commonly used EO benchmarks, TreeSatAI \cite{ahlswede2022treesatai} {and PASTIS-R \cite{garnot2022multi}},
whose focus on crop type mapping and forestry differs from the land cover analysis of FLAIR. We added new modalities to these datasets to reach three distinct data types: VHR aerial images, optical time series, and SAR time series.
See \cref{fig:data} for an illustration, and \cref{sec:exp:datasets} for more details on how we extended these datasets.

\begin{figure}[t]
    \centering
    \input{figures/pipeline}
    \caption{{\bf OmniSat Architecture.} We illustrate OmniSat for $M=3$ modalities, and a tile split into $P=4$ patches. 
    The $M\times P$ input tokens $x^\bM_\bP$ are encoded by $M$ modality-specific encoders $\Enc^\bM$, yielding the token representations $f^\bM_\bP$. The module $\comb$ combines them into multimodal patch representations $f^\star_\bP$.
    The token embeddings $f^\bM_\bP$ are supervised by a contrastive cross-modal objective. We also use a reconstruction objective: the masked multimodal representations $f^\star_\bP$ are decoded by modality-specific networks $\Dec^\bM$ to reconstruct their corresponding inputs in $x^\bM_\bP$.}
    \label{fig:pipeline}
\end{figure}

\section{Method}
We consider a tile $x$ observed through a set $\bM$ of $M$ distinct sensors or modalities. 
The goal of the OmniSat model is to learn in a self-supervised fashion to combine all modalities $\bM$ into a multimodal representation $f^\star$.
We first provide details about OmniSat's architecture in \cref{sec:archi}. We then explain our our training scheme, which consists of a cross-modal contrastive objective (\cref{sec:contrast}) and a multimodal masked encoding task (\cref{sec:mae}). Finally, we present the implementation details in \cref{sec:implem}. The overall method is represented in \cref{fig:pipeline}. 
%
\subsection{Architecture }
\label{sec:archi}
This section presents the tokenization process, the structures of the encoder and decoder for each modality, and the architecture of the modality combiner network.

\parag{\bf Multimodal Tokenization.}
All available modalities are spatially aligned through georeferencing.
This allows us to divide the tile $x$ into a set $\bP$ of $P$ non-overlapping patches consistently across all modalities: 
$x^\bM_p=\{x^m_p\}_{m\in \bM}$
corresponds to $M$ distinct views of the same patch $p$ with different modalities.
Each modality $m$ takes its values in a space $\Omega^m$ such that $x^m_p \in \Omega^m$.
We index tokens with pairs $(m,p)$, defined for each modality $m$ and patch $p$, for a total of $M\times P$ tokens. 

Time series from Sentinel satellites may experience registration errors spanning several meters, complicating their precise alignment with high-resolution imagery. However, using temporal sequences of satellite data mitigates these errors as aggregation over time tends to balance out misalignments.

\parag{\bf Encoder-Decoder for Images.}
We split image tiles split into small square patches: ${\Omega^\text{img}=\bR^{C \times W \times W}}$ with $W$ the size of the patches in pixels and $C$ the number of channels. As shown in \cref{fig:implem:a}, we encode these inputs with a sequence of convolutions and max-pool layers until the spatial dimension is fully collapsed. Decoding involves a symmetric sequence of convolutions and un-pooling layers. Contrary to existing masked auto-encoders, we pass the pooling indices from the encoder's max-pooling to the decoder's un-pooling in the manner of SegNet~\cite{badrinarayanan2017segnet}. This dispenses the encoder from learning the intra-patch spatial configuration. This allows the image encoder to focus on the radiometric information, which may be more relevant depending on the application.

\begin{figure*}[t]
    \centering
    \begin{tabular}{c@{}c@{}c}
   \begin{subfigure}[t]{0.3\linewidth}\centering
     \input{figures/image}
    \caption{\scriptsize{\bf Image \textcolor{ENCCOLOR}{Encoder} and \textcolor{DECCOLOR}{Decoder}.}}
    \label{fig:implem:a}
     \end{subfigure}
     &  
    \begin{subfigure}[t]{0.3\linewidth}\centering
     \input{figures/temporal}
    \caption{\scriptsize{\bf Temporal \textcolor{ENCCOLOR}{Encoder} and \textcolor{DECCOLOR}{Decoder}.}}
    \label{fig:implem:b}
     \end{subfigure}
     & 
     \begin{subfigure}[t]{0.4\linewidth}\centering
     \begin{tikzpicture}
        \node[minimum width=3mm, minimum height=3mm, draw, very thick, fill=MODONECOLOR, anchor=center] at (0,+0) (f11) {};
        \node[minimum width=3mm, minimum height=3mm, draw, very thick, fill=MODONECOLOR, anchor=center] at (0.5,+0) (f12) {};
        \node[minimum width=3mm, minimum height=3mm, draw, very thick, fill=MODTWOCOLOR, anchor=center] at (1,+0) (f21) {};
        \node[minimum width=3mm, minimum height=3mm, draw, very thick, fill=MODTWOCOLOR, anchor=center] at (1.5,+0) (f22) {};
        \node[minimum width=3mm, minimum height=3mm, draw, very thick, fill=MODTHREECOLOR, anchor=center] at (2,+0) (f31) {};
        \node[minimum width=3mm, minimum height=3mm, draw, very thick, fill=MODTHREECOLOR, anchor=center] at (2.5,+0) (f32) {};
      
        \node[minimum width=3mm, minimum height=3mm, draw, very thick, fill=MODONECOLOR, anchor=center] at (0,-2) (f112) {};
        \node[minimum width=3mm, minimum height=3mm, draw, very thick, fill=MODONECOLOR, anchor=center] at (0.5,-2) (f122) {};
        \node[minimum width=3mm, minimum height=3mm, draw, very thick, fill=MODTWOCOLOR, anchor=center] at (1,-2) (f212) {};
        \node[minimum width=3mm, minimum height=3mm, draw, very thick, fill=MODTWOCOLOR, anchor=center] at (1.5,-2) (f222) {};
        \node[minimum width=3mm, minimum height=3mm, draw, very thick, fill=MODTHREECOLOR, anchor=center] at (2,-2) (f312) {};
        \node[minimum width=3mm, minimum height=3mm, draw, very thick, fill=MODTHREECOLOR, anchor=center] at (2.5,-2) (f322) {};







         \node[rectangle, minimum width=28mm, minimum height=7mm, draw, very thick, fill=LOSSCOLOR, anchor=center] (self) at (1.25,-1) {self-attention};

        
        \node[minimum width=3mm, minimum height=3mm, draw, very thick, fill=LOSSCOLOR, anchor=center] at (-1,-2.75) (fs1) {};
        \node[minimum width=3mm, minimum height=3mm, draw, very thick, fill=LOSSCOLOR, anchor=center] at (-1,-3.25) (fs2) {};


        \node[minimum width=3mm, minimum height=3mm, draw, very thick, fill=MODALLCOLOR, anchor=center] at (1,-4) (fs12) {};
        \node[minimum width=3mm, minimum height=3mm, draw, very thick, fill=MODALLCOLOR, anchor=center] at (1.5,-4) (fs22) {};

        \node[rectangle, minimum width=28mm, minimum height=7mm, draw, very thick, fill=LOSSCOLOR, anchor=center] (cross) at (1.25,-3) {cross-attention};

        %
        \node[text width=1.2cm, align=center] (pos) at (-0.7,-1.6) {\scriptsize position encoding};
        \draw[very thick, ->] (pos.south) ++ (0.0,0) -- ++ (0,-1);
        \draw[very thick, ->] (pos.north) ++ (0.0,0) -- ++ (0,1.00) -- ++ (0.3, 0);

        \draw [very thick, ->] (1.25,-0.15) -- (self) -- (1.25 ,-1.85);
    
        \draw [very thick, ->] -- (1.25,-2.15) -- (cross) -- (1.25,-3.85);
        \draw [very thick, ->] (-0.85,42 |- cross) -- (cross);

        \node [right = 0.0cm of f32, anchor=west]  {$f^\bM_\bP$};
        \node [right = 0.0cm of f322, anchor=west] {$g^\bM_\bP$};
        \node [right = 0.0cm of fs22, anchor=west] {$f^\star_\bP$};
        \node [below right = 0.2cm and -0.2cm of fs2, anchor=west] {$f^\text{comb}_\bP$};

\end{tikzpicture}
    \caption{\scriptsize{\bf Modality Combining Network.}}
    \label{fig:implem:c}
     \end{subfigure}
\end{tabular}
    \caption{{\bf OmniSat Architecture.} 
    OmniSat is composed of dedicated patch encoders for image (\subref{fig:implem:a}) and time series \subref{fig:implem:b}, here represented for a length of $L=4$ time stamps. The modality combining module $\comb$ is depicted in (\subref{fig:implem:c}) with $P=2$ and $M=3$. {Elements colored in {orange} are learned networks or parameters.}}
    \label{fig:enter-label}
\end{figure*}

\parag{\bf Encoder-Decoder for Time Series.}
Each temporal patch is represented by $L$ sequential observations with $C$ channels:  $\Omega^\text{TS}=\bR^{C \times L}$, each associated with a time stamp.
We encode the temporal patches using a Lightweight Temporal Attention Encoder (LTAE) model  \cite{garnot2020lightweight}, an efficient network for geospatial time series processing.
We decode vector representations into time series by repeating the vector $L$ times across the temporal dimension, adding a temporal encoding for each time step, and using an MLP to map the results to size $C$. See \cref{fig:implem:b} for an illustration.

Optical time series are notoriously affected by clouds \cite{sudmanns2019assessing}. This may affect the validity of the reconstruction task: the decoder cannot know which observations are cloudy, making the reconstruction objective unpredictable. To circumvent this issue, we use the temporal attention maps of the encoder's LTAE to select dates to reconstruct: cloudless observations are more informative and should have a higher attention score \cite{russwurm2020self}. We only consider in the reconstruction loss $\Lmae$ the top $25\%$ dates in terms of the LTAE's attention maps.

\parag{\bf Modality Combining Network.} 
The modality combining network $\comb$, represented in \cref{fig:implem:c}, takes the $M \times P$ token embeddings ${f}^\bM_\bP$, some of whom can potentially be masked. 
We equip each token with a Euclidean relative positional encoding \cite{wu2021rethinking},  calculated based on their patch's position $\{r(p,q) \mid (p,q)\in \bP^2\}$, allowing each token to selectively consider its spatial surroundings. As most EO data are captured from above (satellite or aerial), their distribution is invariant by horizontal translation, making this choice of encoding preferable to an absolute position encoding.

The modality combining module $\comb$ starts with a series of $B$ residual self-attention blocks connecting all tokens across modality. We then perform cross-attention between the resulting token embeddings $g^\bM_\bP \in \bR^{d \times M \times P}$ and $P$ copies $f^\text{comb}_\bP$ of a modality combining token $f^\text{comb} \in \bR^d$ learned as a free parameter. Each copy of $f^\text{comb}_p$ is spatially located at the patch $p$ for the relative positional encoding $r$. The module $\comb$ outputs $P$ multimodal encodings ${f}^\star_\bP$ combining all available modalities for each patch:
\begin{align}
    g^\bM_\bP & = \text{self-attention}
    \left(
    {f}^{\bM}_{\bP}; r\right) \\
    {f}^\star_\bP & =  \text{cross-attention}
    \left(f^\text{comb}_\bP, g^\bM_\bP ;r
    \right)~.
\end{align}
\subsection{Contrastive Objective}
\label{sec:contrast}
We denote by $f^m_p$ the $d$-dimensional encodings of the input patch $x^m_p$ given by their dedicated encoders. We propose to supervise the embeddings $f^m_p$ with a contrastive objective encouraging spatial consistency \emph{across modalities}. Indeed, while each modality captures distinct characteristics of  $p$, all encodings $f^m_p$ share the same latent variable: the semantic content of the patch.

In practice, we want $f^m_p$ to be closer to $f^n_p$ for $n\neq m$, than to $f^n_q$ for other patches $q\neq p$. We define $\batch$ as the set of patches within the current batch of observations.
We adapt the classic InfoNCE loss  \cite{oord2018representation} to our setting with two main differences, illustrated in \cref{fig:contrast}.
(i) Each token $(m,p)$ has $M-1$ positive matches: the tokens corresponding to the same patch $p$ but viewed in another modality $n\neq m$;
and (ii) as EO observations are generally spatially regular, nearby patches may be visually indistinguishable. Therefore, we exclude from the negative matches of $(m,p)$ all tokens in modality $m$ and which are too close to $p$. To this end, we remove the set $T(m,p)$ of tokens with modality $m$ and whose patches are in the same tile as $p$.
Our loss function $\Lcon$ is defined as such:
\begin{align}
    \Lcon = \frac1{M\vert \batch \vert}
    \sum_{(m,p) \in \bM \times \batch}
    \log 
    \left(
        \frac
        {
        \sum_{n \neq m}
            \exp(\langle f^m_p, f^n_p\rangle / \gamma)
        }
        {
        \sum_{(n,q)\in \bM\times \batch \setminus T(m,p)}
            \exp(\langle f^m_p, f^n_q\rangle / \gamma)
        }
    \right)~,
\end{align}
with $\gamma$ a temperature parameter,  and 
$\langle \cdot,\cdot\rangle$ the scalar product in $\bR^d$.
This function, specifically designed for geospatial data, allows us to contrast individual patches across modalities, which is not typically feasible for natural images.
However, as the contrastive objective aligns multimodal representations, the patch encoders  may be encouraged to overlook the distinct attributes of their respective modality. Instead, they may focus only on features shared by all modalities, \ie, their \emph{common denominator}. To ensure that encoders also capture modality-specific information, we incorporate a reconstruction objective, detailed in \cref{sec:mae}.

 \begin{figure}[t]
     \centering
     \begin{minipage}[c]{0.45\textwidth}
    \newcolumntype{C}[1]{>{\centering\arraybackslash}p{#1}}

\resizebox{\linewidth}{!}{
\begin{tabular}{lll C{3.8mm}C{3.8mm} C{3.8mm}C{3.8mm} C{3.8mm}C{3.8mm}  C{3.8mm}C{3.8mm} C{3.8mm}C{3.8mm} C{3.8mm}C{3.8mm}}
&&&\multicolumn{6}{c}{$\tile_1$} & \multicolumn{6}{c}{$\tile_2$} \\
\cmidrule(r{4pt}){4-9}\cmidrule(r{4pt}){10-15}
&&&\multicolumn{2}{c}{$m_1$}  & \multicolumn{2}{c}{$m_2$} & \multicolumn{2}{c}{$m_3$}
&\multicolumn{2}{c}{$m_1$}  & \multicolumn{2}{c}{$m_2$} & \multicolumn{2}{c}{$m_3$} \\
\cmidrule(r{4pt}){4-5}\cmidrule(r{4pt}){6-7}
\cmidrule(r{4pt}){8-9}\cmidrule(r{4pt}){10-11}
\cmidrule(r{4pt}){12-13}\cmidrule(r{4pt}){14-15} 
&&&$p_1$ & $p_2$ & $p_1$ & $p_2$ & $p_1$ & $p_2$ & $q_1$ & $q_2$ & $q_1$ & $q_2$ & $q_1$ & $q_2$ \\
\multirow{6}{*}{\raisebox{+.6cm}{\rotatebox{90}{$\tile_1$}} {\vrule height 2cm}}&\multirow{2}{*}{\raisebox{+.2cm}{$m_1$} \vrule height .6cm} & \vphantom{\vrule height .35cm}$p_1$ &
\cellcolor{black!30}$\mkern-1.5mu$o& \cellcolor{black!30}$\mkern-1.5mu$o & \cellcolor{green}$\mkern-1.5mu$+ & \cellcolor{red!50}- & \cellcolor{green}$\mkern-1.5mu$+ & \cellcolor{red!50}- & \cellcolor{red!50}- & \cellcolor{red!50}- & \cellcolor{red!50}- & \cellcolor{red!50}- & \cellcolor{red!50}- & \cellcolor{red!50}- \\ 
&& \vphantom{\vrule height .35cm}$p_2$&
\cellcolor{black!30}$\mkern-1.5mu$o & \cellcolor{black!30}$\mkern-1.5mu$o & \cellcolor{red!50}- & \cellcolor{green}$\mkern-1.5mu$+ & \cellcolor{red!50}- & \cellcolor{green}$\mkern-1.5mu$+ & \cellcolor{red!50}- & \cellcolor{red!50}- & \cellcolor{red!50}- & \cellcolor{red!50}- & \cellcolor{red!50}- & \cellcolor{red!50}- \\
&\multirow{2}{*}{\raisebox{+.2cm}{$m_2$} \vrule height .6cm} & \vphantom{\vrule height .35cm}$p_1$ &
\cellcolor{green}$\mkern-1.5mu$+ & \cellcolor{red!50}- & \cellcolor{black!30}$\mkern-1.5mu$o & \cellcolor{black!30}$\mkern-1.5mu$o & \cellcolor{green}$\mkern-1.5mu$+ & \cellcolor{red!50}- & \cellcolor{red!50}- & \cellcolor{red!50}- & \cellcolor{red!50}- & \cellcolor{red!50}- & \cellcolor{red!50}- & \cellcolor{red!50}- \\
& & \vphantom{\vrule height .35cm}$p_2$&
\cellcolor{red!50}- & \cellcolor{green}$\mkern-1.5mu$+ & \cellcolor{black!30}$\mkern-1.5mu$o & \cellcolor{black!30}$\mkern-1.5mu$o & \cellcolor{red!50}- & \cellcolor{green}$\mkern-1.5mu$+ & \cellcolor{red!50}- & \cellcolor{red!50}- & \cellcolor{red!50}- & \cellcolor{red!50}- & \cellcolor{red!50}- & \cellcolor{red!50}- \\
&\multirow{2}{*}{\raisebox{+.2cm}{$m_3$} \vrule height .6cm} & \vphantom{\vrule height .35cm}$p_1$&
\cellcolor{green}$\mkern-1.5mu$+ & \cellcolor{red!50}- & \cellcolor{green}$\mkern-1.5mu$+ & \cellcolor{red!50}- & \cellcolor{black!30}$\mkern-1.5mu$o & \cellcolor{black!30}$\mkern-1.5mu$o & \cellcolor{red!50}- & \cellcolor{red!50}- & \cellcolor{red!50}- & \cellcolor{red!50}- & \cellcolor{red!50}- & \cellcolor{red!50}- \\
& & \vphantom{\vrule height .35cm}$p_2$&
\cellcolor{red!50}- & \cellcolor{green}$\mkern-1.5mu$+ & \cellcolor{red!50}- & \cellcolor{green}$\mkern-1.5mu$+ & \cellcolor{black!30}$\mkern-1.5mu$o & \cellcolor{black!30}$\mkern-1.5mu$o & \cellcolor{red!50}- & \cellcolor{red!50}- & \cellcolor{red!50}- & \cellcolor{red!50}- & \cellcolor{red!50}- & \cellcolor{red!50}- \\
\multirow{6}{*}{\raisebox{+.6cm}{\rotatebox{90}{$\tile_1$}} {\vrule height 2cm}}&\multirow{2}{*}{\raisebox{+.2cm}{$m_1$} \vrule height .6cm}&\vphantom{\vrule height .35cm}$q_1$ &
 \cellcolor{red!50}- & \cellcolor{red!50}- & \cellcolor{red!50}- & \cellcolor{red!50}- & \cellcolor{red!50}- & \cellcolor{red!50}- & \cellcolor{black!30}$\mkern-1.5mu$o & \cellcolor{black!30}$\mkern-1.5mu$o & \cellcolor{green}$\mkern-1.5mu$+ & \cellcolor{red!50}- & \cellcolor{green}$\mkern-1.5mu$+ & \cellcolor{red!50} -\\ 
&& \vphantom{\vrule height .35cm}$q_2$&
 \cellcolor{red!50}- & \cellcolor{red!50}- & \cellcolor{red!50}- & \cellcolor{red!50}- & \cellcolor{red!50}- & \cellcolor{red!50}- & \cellcolor{black!30}$\mkern-1.5mu$o & \cellcolor{black!30}$\mkern-1.5mu$o & \cellcolor{red!50}- & \cellcolor{green}$\mkern-1.5mu$+ & \cellcolor{red!50}- & \cellcolor{green}$\mkern-1.5mu$+ \\
&\multirow{2}{*}{\raisebox{+.2cm}{$m_2$} \vrule height .6cm} & \vphantom{\vrule height .35cm}$q_1$&
 \cellcolor{red!50}- & \cellcolor{red!50}- & \cellcolor{red!50}- & \cellcolor{red!50}- & \cellcolor{red!50}- & \cellcolor{red!50}- &  \cellcolor{green}$\mkern-1.5mu$+ & \cellcolor{red!50}- & \cellcolor{black!30}$\mkern-1.5mu$o & \cellcolor{black!30}$\mkern-1.5mu$o & \cellcolor{green}$\mkern-1.5mu$+ & \cellcolor{red!50}- \\
& & \vphantom{\vrule height .35cm}$q_2$&
 \cellcolor{red!50}- & \cellcolor{red!50}- & \cellcolor{red!50}- & \cellcolor{red!50}- & \cellcolor{red!50}- & \cellcolor{red!50}- & \cellcolor{red!50}- & \cellcolor{green}$\mkern-1.5mu$+ & \cellcolor{black!30}$\mkern-1.5mu$o & \cellcolor{black!30}$\mkern-1.5mu$o & \cellcolor{red!50}- & \cellcolor{green}$\mkern-1.5mu$+ \\
&\multirow{2}{*}{\raisebox{+.2cm}{$m_3$} \vrule height .6cm} & \vphantom{\vrule height .35cm}$q_1$&
 \cellcolor{red!50}- & \cellcolor{red!50}- & \cellcolor{red!50}- & \cellcolor{red!50}- & \cellcolor{red!50}- & \cellcolor{red!50}- & \cellcolor{green}$\mkern-1.5mu$+ & \cellcolor{red!50}- & \cellcolor{green}$\mkern-1.5mu$+ & \cellcolor{red!50}- & \cellcolor{black!30}$\mkern-1.5mu$o & \cellcolor{black!30}$\mkern-1.5mu$o\\
& & \vphantom{\vrule height .35cm}$q_2$&
 \cellcolor{red!50}- & \cellcolor{red!50}- & \cellcolor{red!50}- & \cellcolor{red!50}- & \cellcolor{red!50}- & \cellcolor{red!50}- & \cellcolor{red!50}- & \cellcolor{green}$\mkern-1.5mu$+ & \cellcolor{red!50}- & \cellcolor{green}$\mkern-1.5mu$+ & \cellcolor{black!30}$\mkern-1.5mu$o & \cellcolor{black!30}$\mkern-1.5mu$o\\
\end{tabular}
}
    \end{minipage}
    \hfill
    \begin{tabular}{c}
    \begin{minipage}[c]{0.5\textwidth}
    \caption{{\bf Contrastive Loss.} We represent the token matching matrix for two tiles $\tile_1$ and $\tile_2$  viewed across $3$ modalities $m_1$, $m_2$, and $m_3$. $\tile_1$ is composed of the patches $p_1$ and $p_2$, while $\tile_2$ comprises $q_1$ and $q_2$. In contrast to classic approaches which ignore the diagonal and assign each sample with a single positive match, our loss defines operates at the patch level, considers multiple positives per token, and excludes tokens in a block-diagonal fashion.
    }
    \label{fig:contrast}
    \end{minipage}
    \\
     \begin{minipage}[c]{0.35\textwidth}
    \centering 
    \begin{tabular}{r@{\;\;}l}
    \tikz[baseline]  \node [fill=green, minimum width=4mm, minimum height=4mm,anchor=base] {+};&\small positive match
    \\
   \tikz[baseline]  \node [fill=red!50, minimum width=4mm, minimum height=4mm,anchor=base] {-\vphantom{+}};&\small negative match
   \\
   \tikz[baseline]  \node [fill=black!30, minimum width=4mm, minimum height=4mm,anchor=base] {o\vphantom{+}};&\small ignored
    \end{tabular}
    \end{minipage}
    \end{tabular}
 \end{figure}

\subsection{Multimodal Reconstruction Objective}
\label{sec:mae}
During training, we mask a fraction of tokens $\mask \subset \bM\times \bP$ and replace their embeddings with a learned vector $f^\text{mask} \in \bR^d$. Note that the masking can differ across modalities, and some patches may be entirely masked.
All tokens are then processed by the modality combining network $\comb$, which outputs $P$ multimodal embeddings $f^\star_{\bP}$:
\begin{align}
    f_\bP^\star= \comb
    \left(
    \{f^m_p\}_{(m,p) \not\in \mask}
    \cup
    \{f^\text{mask}\}_{(m,p) \in \mask}
    \right)~.
\end{align}
To encourage the patch embeddings $f^\star_{\bP}$ to capture information from all modalities, we build a multimodal reconstruction objective.
We denote by $\Dec^m:\bR^d \mapsto \Omega^m$ the  dedicated decoder of each modality $m$ and write the reconstruction loss as:
\begin{align}
    \Lmae =
    \frac1{\vert \mask \vert}
    \sum_{(m,p) \in \mask} 
    \frac1{\text{dim}(\Omega^m)}
    \left\Vert 
        \Dec^m(f^\star_p) - x^m_p
    \right\Vert^2~,
\end{align}
with $\text{dim}(\Omega^m)$ the dimension of $\Omega^m$.
The total loss is the sum of $\Lmae$ and $\Lcon$.%

\subsection{Implementation Details}
\label{sec:implem}
We detail here the specific parameters chosen in all our experiments. 

\parag{\bf Tokenization.} 
 We split each tile along a regular spatial grid to produce a set of non-overlapping patches $\bP$ consistent across all modalities.
{For TreeSat and FLAIR, we use a $10\times10$ m grid, meaning that the VHR input tokens are small image patches of size $50\times 50$ with $0.2$ m per pixel. The patches of Sentinel observations with a resolution of $10$m are single-pixel temporal sequences of spectral measurements. For PASTIS-HD, we use a $40\times40$ m grid, meaning that the VHR patches are of size $40\times 40$ with $1.0$ m per pixel. The patches of Sentinel observations \cite{Sentinel-2-paper} are $4\times4$ image time series which we spatially flatten before encoding.}

\parag{\bf Hyperparameters.} To show the versatility of OmniSat, we use the same configuration throughout all experiments. The embedding size is $d=256$, resulting in image encoders and decoders with $3.6$M and $1.8$M parameters, $403$k and $96$k for optical time series, and $402$k and $95$k for radar time series. The modality combiner module is composed of $B=6$ residual self-attention blocks and a single cross-attention block, for a total of $3.6$M parameters. We train our model on 3 A6000 GPUs with a batch size of $128$ multimodal tiles per GPU and set the contrastive temperature $\gamma$ to $0.1$. We train our model with the ADAM optimizer \cite{kingma2014adam}, with a learning rate of $10^{-4}$ for pretraining and $2\times 10^{-5}$ for fine-tuning, and a ReduceLROnPlateau scheduler \cite{reduce} with a patience of $10$ epochs and a decay rate of $0.1$. 
When re-implementing competing methods, we use the hyperparameters of their open-source repository.
\section{Experiments}

We evaluate OmniSat's performance across three multimodal datasets, including two new datasets introduced in this work, and presented in \cref{sec:exp:datasets}. We outline our experimental protocol and our adaptation of competing methods in \cref{sec:exp:baselines}. We then present  our quantitative results and analysis in \cref{sec:exp:exp} 
and conduct an ablation study in \cref{sec:exp:ablation}.
\subsection{Datasets}
\label{sec:exp:datasets}
Weconsider three multimodal datasets: FLAIR \cite{garioud2023flair}, and the augmented TreeSatAI-TS \cite{ahlswede2022treesatai} and PASTIS-HD \cite{garnot2021panoptic,garnot2022multi}. See \cref{fig:data} for an illustration of these two last datasets.

\parag{\bf \bf{TreeSatAI-TS}:} 
TreeSatAI \cite{ahlswede2022treesatai} is a multimodal dataset for tree species identification, containing 50,381 tiles of $60\times60$ m with multi-label annotations for $20$ classes and all taken in Germany. Each tile is associated with a very high resolution RGB and near-infrared (NIR) image ($0.2$ m pixel resolution), a single Sentinel-2 multi-spectral image ($10$ m per pixel resolution, 10 bands), and a single Sentinel-1 radar image ($10$ m per pixel resolution, 3 bands: two polarization channels and their ratio).

Motivated by the fact that fine-grained vegetation discrimination relies heavily on temporal dynamics information \cite{vrieling2018vegetation}, we introduce TreeSatAI-TS\footnote{The dataset is available at \href{https://huggingface.co/datasets/IGNF/TreeSatAI-Time-Series}{https://huggingface.co/datasets/IGNF/TreeSatAI-Time-Series}.}. This extended version uses open-source data to add Sentinel-1 and Sentinel-2 time series to each tile, spanning the closest available year to the VHR observation for Sentinel-2. Note that due to the weather patterns and position of the area of interest with respect to Sentinel-2's orbit, the optical time series is particularly irregular and occluded, with up to $50$\% of acquisitions being non-exploitable. Despite this challenge, we included the raw observations without pre-processing, whereas TreeSatAI's single-date images have been manually selected.

\parag{\bf \textbf{PASTIS-HD}:}
{The PASTIS dataset \cite{garnot2021panoptic}, is designed for semantic and panoptic segmentation of agricultural parcels using Sentinel-2 time series and covers $18$ crop types across $2,433$ image time series with dimensions of $1280\times1280$ m. Each series contains between $38$ and $61$ observations with $10$ spectral bands. PASTIS-R \cite{garnot2022multi} adds the corresponding Sentinel-1 radar time series. {We only used the ascendent time series of Sentinel-1 for our training and evaluation, for a total of 169,587 radar images with three bands.}

{To enhance the spatial resolution and utility of PASTIS, we introduce PASTIS-HD\footnote{The dataset is available at \href{https://huggingface.co/datasets/IGNF/PASTIS-HD}{https://huggingface.co/datasets/IGNF/PASTIS-HD}.
}, which integrates contemporary VHR satellite images (SPOT 6-7 \cite{spot}). We apply orthorectification and pansharpening, resample the resulting images to a $1$m resolution, and finally convert them to 8 bits.}
We follow the protocol of Irvin \etal  \cite{irvin2023usat} to use the dense annotations for a multi-label classification task: each patch is associated with the labels of all of its pixels. This conversion allows us to evaluate all methods in the same setting and configuration as TreeSatAI.

\parag{\bf FLAIR.} 
The FLAIR dataset \cite{garioud2023flair} combines VHR aerial images with time series data. It comprises 77,762 aerial tiles ($512\times512$ pixels, $0.2$ m resolution) with five channels (RGB, near-infrared, and a normalized digital surface model) taken in France, alongside corresponding Sentinel-2 time series ($10$ m resolution, $10$ spectral bands, $20$ to $114$ observations per year). 
{We apply the same processing as PASTIS to use the dense annotation for a multi-label classification task.}

\subsection{Experimental Setting}
\label{sec:exp:baselines}
This section details our experimental protocol and our adaption of competing algorithms.

\parag{\bf Evaluation Protocol.} All experiments follow a similar setting:
\begin{itemize}
    \item {\bf Pre-training (optional).} Methods that support self-supervised pre-training (OmniSat, SatMAE \cite{cong2022satmae}, ScaleMAE \cite{reed2023scale}, CROMA \cite{fuller2023croma}) are pre-trained for up to $250$ epochs on the entire training set without access to labels.
    \item {\bf Fine-Tuning.} We propose two settings for fine-tuning:
    \begin{itemize}
    \item[$\bullet$] {\bf Fully Supervised Fine-Tuning.} We train the resulting models using all the labels in the training set. 
    \item[$\bullet$]  {\bf Semi-Supervised Fine-Tuning.} We use a portion of $10$\% or $20\%$ of the training set, stratified by the distribution of classes, to fine-tune the models. For models without pre-training, this corresponds to supervision in the low-data regime.
    \end{itemize}
    \item {\bf Unimodal and Multimodal Evaluation.} We evaluate all methods using each available modality independently and combining all supported modalities. 
\end{itemize}


\begin{table}[t]
    \centering
    \setlength{\fboxsep}{1pt}
\caption{{\bf Performance on TreeSatAI-TS.} 
We report the weighted F1 for multi-label tree species classification on TreeSatAI (TSAI) and our extended TreeSatAI-TS (TSAI-TS) dataset when fine-tuning with $10$\% and $100$\% of training labels. The first line of the table is the modality used for evaluation.
We distinguish methods that are \textbf{best for one modality} within a dataset, \textbf{\underline{best in a dataset}} across all modalities, and the \textbf{\fbox{best overall}}~performance. 
$^\star$: late feature fusion with a ResNet pre-trained on ImageNet. \globe: \small Foundation model trained on extensive external data.\\ 
$\protect\Dag$\small: model evaluated on this dataset for the first time.\vspace{-1mm}}
\label{tab:tsai}
\resizebox{\linewidth}{!}{
\begin{tabular}{ll cc cc cc cc}
\toprule
\multirow{2}{*}{Model} & pre- &  \multicolumn{2}{c}{~\hspace{6mm} \bf All \hspace{6mm}~ } & \multicolumn{2}{c}{ ~\hspace{1mm} Sentinel-1 \hspace{1mm}~ } & \multicolumn{2}{c}{~\hspace{1mm} Sentinel-2 \hspace{1mm}~} & \multicolumn{2}{c}{~\hspace{1mm} VHR Image \hspace{1mm}~} \\\cmidrule(r{4pt}){3-4} \cmidrule(r{4pt}){5-6} \cmidrule(r{4pt}){7-8} \cmidrule(r{4pt}){9-10}
& {training\hspace{5mm}~}& 10\% & 100\% & 10\% & 100\% & 10\% & 100\% & 10\% & 100\%\\
\midrule
\multicolumn{10}{c}{Evaluated on TreeSatAI: single date for Sentinel-1 and Sentinel-2}\\\midrule
$\Dag$ PSE \cite{garnot2020satellite} & None& 47.2$^\star$ & 68.1$^\star$  & 11.5 & 14.6  &  \bf 48.5 & \bf  58.3 & - & - \\
\rowcolor{black!5} $\Dag$ ViT \cite{dosovitskiy2020image}& None& 42.7\deadstar & 57.1\deadstar & 8.7 & 17.5 &  39.8 & 57.3 & 36.7& 51.7\\
 MLP\cite{ahlswede2022treesatai} & None &  42.6$^\star$ & 71.5$^\star$& 3.4 & 10.1 & 22.1 & 52.0 & - & - \\
 \rowcolor{black!5} ResNet \cite{ahlswede2022treesatai} & ImageNet & - & - & - & - & - & -& 58.8 & \bf 70.1  \\
LightGBM \cite{ahlswede2022treesatai} & ImageNet &  - & 54.3$^\star$  &  - & 11.9 & - & 48.2 &  - & 44.0\\
\rowcolor{black!5} PRESTO \cite{tseng2023lightweight} & \;\;\globe & - & - &  - & \bf 19.8 & - & 46.3 & - & - \\
$\Dag$ DOFA \cite{xiong2024dofa} & \;\;\globe & \bestB{59.5}\deadstar & \bestB{71.6}\deadstar & \bf 11.6 & 19.3 & 48.2 & 57.0 & 51.6 & 67.5 
\\\greyrule
  \rowcolor{black!5}MOSAIKS \cite{corley2023revisiting,rolf2021generalizable} & TSAI & - & -& - &- & -& 56.0 & - & -
\\
 $\Dag$ CROMA \cite{fuller2023croma} &TSAI & 49.6\deadstar & 61.0\deadstar & 10.1 & 12.7 & 47.8 & 55.7 & - & -\\
 \rowcolor{black!5} $\Dag$ SatMAE  \cite{cong2022satmae}& TSAI & 46.1\deadstar  & 61.5\deadstar & - & - & 40.3 & 49.7 & 44.1 & 61.4\\
$\Dag$ ScaleMAE \cite{reed2023scale} &TSAI & 47.6\deadstar & 62.5\deadstar  & - & - & 46.7 & 55.2 & 46.9 & 63.6
\\\greyrule
 \rowcolor{black!5} \bf OmniSat (ours) &TSAI  & {56.2}\deadstar & 70.4\deadstar & 5.3 & 6.4 & 48.4 & 57.1 & 52.8 & 68.9\\
\midrule
\multicolumn{10}{c}{Evaluated on TreeSatAI-TS: Sentinel-1 and Sentinel-2 time series spanning one year}\\\midrule
 $\Dag$  PSE+LTAE  \cite{garnot2020satellite} & None& 59.4$^\star$ & 71.2$^\star$ &  42.6 & 52.4 & 44.0 & 57.2 & - & -\\ 
\rowcolor{black!5}   $\Dag$ UT\&T \cite{garioud2023flair} & ImageNet & 43.8\deadstar & 56.7\deadstar & 42.3 & 55.2 & 41.5 & 57.0 & 44.3 & 55.9 \\ 
 $\Dag$ DOFA \cite{xiong2024dofa} & \;\;\globe & 41.8\deadstar & 71.3\deadstar & 0.0 & 0.0 & 25.0 & 39.4 & 51.6 & 67.5 
\\ \greyrule
 \rowcolor{black!5} $\Dag$ Scale-MAE \cite{reed2023scale} &TSAI-TS& 44.1\deadstar  & 60.4\deadstar  & - & - & 11.0 & 31.5 & 46.9 & 63.6\\\greyrule
  \bf  OmniSat (ours) & None& 52.2\deadstar & 73.3\deadstar & 31.6 & 55.9 & 33.9 & 49.7 & 51.4 & \bestA{71.0} \\ 
 \rowcolor{black!5}   \bf OmniSat (ours) & TSAI-TS & \bestC{61.1} & \bestC{74.2} & \bestA{48.2} & \bestA{56.7} & \bestA{51.4} & \bestA{62.9} & \bestA{58.3} & {70.5} \\
\bottomrule
\end{tabular}
}
\end{table}

\parag{\bf Adapting Competing Approaches.} 
We report the performance of several methods taken from the literature on our considered datasets: LightGBM \cite{ahlswede2022treesatai}, PRESTO \cite{tseng2023lightweight}, and MOSAIKS \cite{rolf2021generalizable}. 
However, few existing methods can operate on single- and multi-date data at the same time. To ensure a fair evaluation of competing approaches, we modify various state-of-the-art models to handle a broader combination of modalities. We provide details on these changes in the appendix.

\begin{table}[t]   
\caption{{\bf Performance on  PASTIS-HD.} We report {the macro-averaged F1-score for crop-type multi-class classification} 
 on the PASTIS-HD dataset. We distinguish methods that are \textbf{best for one modality}, \textbf{\underline{best in a dataset}} across all modalities. $^\star$: late feature fusion with a ResNet. $\protect\Dag$: model evaluated on this dataset for the first time.\vspace{-1mm}}
    \label{tab:pastis}
    \centering
   \resizebox{\linewidth}{!}{
\begin{tabular}{ll cc cc cc cc}
\toprule
\multirow{2}{*}{Model} 
& pre- &\multicolumn{2}{c}{~\hspace{6mm} All \hspace{6mm}~ } 
&\multicolumn{2}{c}{ ~\hspace{1mm} Sentinel-1 \hspace{1mm}~ } 
& \multicolumn{2}{c}{~\hspace{1mm} Sentinel-2 \hspace{1mm}~} 
& \multicolumn{2}{c}{~\hspace{1mm} VHR image \hspace{1mm}~} \\\cmidrule(r{4pt}){3-4} \cmidrule(r{4pt}){5-6} \cmidrule(r{4pt}){7-8} \cmidrule(r{4pt}){9-10}
& {trained\hspace{5mm}~}& 20\% & 100\% & 20\% & 100\% & 20\% & 100\% & 20\% & 100\%\\
\midrule
\rowcolor{black!5}$\Dag$ UTAE \cite{garnot2021panoptic, garnot2022multi} & None & 36.8$^\star$ & 46.9$^\star$ & 20.1 & 40.7 & 32.7 & 37.6 & - & - \\
 $\Dag$ ResNet50 \cite{he2016deep} & ImageNet & - &  - & - & - & - & - & \bestA{57.6} & \bestA{59.3} \\
 \rowcolor{black!5} $\Dag$ UT\&T \cite{garioud2023flair} & ImageNet & 54.2\deadstar & 53.5\deadstar & 58.8 & 62.8 & 54.9 & 61.3 & 51.1 & 49.8 \\
 $\Dag$ DOFA \cite{xiong2024dofa} & \;\;\globe & 53.7\deadstar & 55.7\deadstar & 36.7 & 41.5 & 50.8 & 53.4 & 47.9 & 54.8 \\\greyrule
\rowcolor{black!5}
 $\Dag$ Scale-MAE \cite{reed2023scale} & PASTIS-HD & 42.0\deadstar & 42.2\deadstar & - & - & 41.2 & 46.1 & 48.8 & 51.9 \\
 $\Dag$ CROMA \cite{fuller2023croma} & PASTIS-HD & 57.5\deadstar & 60.1\deadstar & 55.3 & 56.1 & 53.0 & 56.7 & - & - \\\greyrule
\rowcolor{black!5} \bf OmniSat (ours) & No & 42.0\deadstar & 59.1\deadstar & 58.2 & 60.2 & 51.7 & 60.1 & 47.3 & 52.8 \\
 \bf OmniSat (ours) & PASTIS-HD & \bestB{62.6}\deadstar & \bestA{69.9}\deadstar & \bestA{60.8} & \bestA{69.0} & \bestA{61.8} & \bestB{70.8} & 54.6 & \bestA{59.3} \\
\bottomrule
\end{tabular}
}
\end{table}

\begin{table}[t]  
\caption{{\bf Performance on  FLAIR.} We report the macro-averaged F1-score for land cover multi-class classification on the FLAIR dataset. We distinguish methods that are \textbf{best for one modality} and \textbf{\underline{best in a dataset}}. $\protect\Dag$: model evaluated on this dataset for the first time.\vspace{-1mm}}
    \label{tab:flair}
    \centering
    \small{
\begin{tabular}{ll cc cc cc}
\toprule
\multirow{2}{*}{Model} & pre- 
& \multicolumn{2}{c}{~\hspace{6mm} All \hspace{6mm}~ }
& \multicolumn{2}{c}{~\hspace{1mm} Sentinel-2 \hspace{1mm}~} 
& \multicolumn{2}{c}{~\hspace{1mm} VHR Image \hspace{1mm}~}  \\\cmidrule(r{4pt}){3-4} \cmidrule(r{4pt}){5-6} \cmidrule(r{4pt}){7-8}
& {trained\hspace{5mm}~}& 10\% & 100\% &10\% & 100\% & 10\% & 100\%\\
\midrule

\rowcolor{black!5}$\Dag$ UT\&T \cite{garioud2023flair} & ImageNet & 44.2 & 48.8 & \bestA{57.4} & 62.0 & 58.9 & 65.5 \\
$\Dag$ DOFA \cite{xiong2024dofa} & \;\;\globe & \bestB{70.6} & \bestB{74.9}& 57.0 & 61.0 &  \bestA{66.8} & \bestA{72.1} \\\greyrule
$\Dag$ ScaleMAE\cite{reed2023scale} & FLAIR  & 63.1 & 70.0 & 52.5 & 61.0 & 61.2 & 67.3 \\ \greyrule
\rowcolor{black!5} \bf OmniSat (ours) & No & 62.5 & 70.0 & 56.1 & \bestA{65.4} & 64.7 & 71.5 \\
\bf OmniSat (ours) & FLAIR & 60.6 & {73.4} & 56.8 & \bestA{65.4} & 65.2 & 71.6 \\
\bottomrule
\end{tabular}
}
\end{table}

\subsection{Numerical Experiments and Analysis}
\label{sec:exp:exp}
In this section, we report our model's performance and efficiency compared to other approaches across the considered datasets and propose our analysis.

\parag{\bf TreeSatAI-TS.} 
\cref{tab:tsai} presents the performance of different models on TreeSatAI and TreeSatAI-TS. We report several key observations:
\begin{itemize}
    \item {\bf Benefit of Time Series.} For the original TreeSatAI dataset with single-date Sentinel-1/2 observations, none of the pre-training schemes significantly improve performance beyond simple baselines such as ResNet, PSE, or MLP, even in a semi-supervised setting. In particular, single-date S1 observations yield low performance for all methods (below $20$ F1-score), emphasizing the need to use the entire time series.
    
    OmniSat exhibits significantly improved results on TreeSatAI-TS, with or without pretraining. Image models struggles to extract meaningful features temporally aggregated temporal observations, while OmniSat learn rich dynamic features.

    The foundation model DOFA \cite{xiong2024dofa}, with 111M parameters and a large closed-source training set, outperforms all models when evaluated on single-date modalities. However, OmniSat reaches higher performances on TreeSatAI-TS with only 10 million parameters, which we attribute to its ability to leverage temporal modalities.
    %
    \item {\bf Benefits of Multimodality.} When using all modalities, OmniSat outperforms all competing methods by a margin of $3$\% F1-score. The multimodal performance of OmniSat and CROMA, which learn to combine data sources, is strictly superior to the F1-score of their best modality by $3.7$\% and $5.3$\% points, respectively. 
    Conversely, the performance of methods that rely on late-fusion  (SatMAE, ScaleMAE, ViT) is comparable to their best modality. This demonstrates the value of learning to combine information from different sources end-to-end. 
    \item {\bf Benefits of Cross-Modal Pre-Training.} 
    With access to all modalities, our self-supervised pre-training improves by $0.9$\% point the F1-score of the model fine-tuned on $100$\% of labels, compared to not pre-training, and $8.9$\% when using only $10$\% of labels. This shows that our pre-training leads to more expressive multimodal features. Interestingly, when performing inference with Sentinel-2 time series alone, the performance increase linked to the pre-training becomes $13.2$\% with $100$\% labels and $17.5$\% with $10$\%. This illustrates that our self-supervised pre-training scheme improves the features learned by each encoder despite not relying on annotated data.
\end{itemize}

\begin{figure}[t]
    \centering
\begin{tikzpicture}
\begin{semilogxaxis}[%
    width=.6\linewidth,
    height=.2\textheight,
    scale only axis,
    ytick={55,60,65,70,75},
    xtick={10000000,50000000,100000000},
    xticklabels={$10^7$,$5\cdot10^7$,$10^8$},
    xmin=10000000,
    xmax=125000000,
    ymin=55,
    ymax=75,
    axis x line*=bottom,
    axis y line*=left,
    xminorgrids=true,
    yminorgrids=true,
    minor grid style={gray!15,line width=.25pt},
    legend pos=south east,
    ylabel={mF1},
    xlabel={Model Size (number of trainable parameters)},
    ylabel style={xshift=-0cm,yshift=-0.4cm},
    clip marker paths=false,
    clip mode=individual,
    ]
]


\node[draw, circle, inner sep=5,draw=none, fill=OMNICOLOR] (omni) at (axis cs:11600000,74.2) {};
\node[draw, circle, inner sep=4.1,draw=none, fill=OMNICOLOR!50] at (axis cs:11600000,74.2) {};
\node [right=0.1cm of omni.east, text=OMNICOLOR] {\bf \underline{OmniSat} (ours)};

\node[draw, circle, inner sep=16.7,draw=none, fill=SATMAECOLOR] (scale) at (axis cs:43300000,61.5) {};
\node[draw, circle, inner sep=14.4,draw=none, fill=SATMAECOLOR!50] at (axis cs:43300000,61.5) {};
\node [above right=0cm and -0.1cm of scale.west, text=SATMAECOLOR!95!black, anchor=east] {\bf SatMAE};

\node[draw, circle, inner sep=16,draw=none, fill=SCALECOLOR, opacity=0.8] (scale) at (axis cs:44000000,62.5) {};
\node[draw, circle, inner sep=13.7,draw=none, fill=SCALECOLOR!50, opacity=0.8] at (axis cs:44000000,62.5) {};
\node [above right=0.5cm and 1.5cm of scale.east, text=SCALECOLOR, anchor=east] {\bf ScaleMAE};

\node[draw, circle, inner sep=2.9,draw=none, fill=TAECOLOR] (tae) at (axis cs:24300000,71.2) {};
\node [below=0.1cm of tae.south, text=TAECOLOR] {\bf PSE+LTAE+ResNet};

\node[draw, circle, inner sep=4.1,draw=none, fill=UTTCOLOR] (tae) at (axis cs:27000000,56.7) {};
\node [right=0.1cm of tae.east, text=UTTCOLOR] {\bf UT\&T};

\node[draw, circle, inner sep=3.8,draw=none, fill=CROMACOLOR] (croma) at (axis cs:50000000,61) {};
\node[draw, circle, inner sep=1.6,draw=none, fill=CROMACOLOR!50] at (axis cs:50000000,61) {};
\node [right=5mm of croma.west, text=CROMACOLOR] {\bf CROMA};

\node[draw, circle, inner sep=5.6,draw=none, fill=DOFACOLOR] (dofa) at (axis cs:111000000,72) {};
\node[draw=none] (dofa) at (axis cs:111000000,72) {\faGlobe};
\node [left=0.3cm of dofa.west, text=DOFACOLOR] {\bf DOFA};

\def\legpos{3000000}

\node[draw=black, circle, inner sep=10, anchor=south] (n1) at (axis cs:\legpos,57.5) {};
\node[draw=black, circle, inner sep=5, anchor=south] (n2) at (axis cs:\legpos,57.5) {};
\draw [-] (n1.north) -- ++ (0.8cm,0cm);
\draw [-] (n2.north) -- ++ (0.8cm,0cm);
\node[draw=none,anchor=north, right=0.8cm of n1.north] (x) {\footnotesize 100h};
\node[draw=none,anchor=west, right=0.8cm of n2.north] (x) {\footnotesize  25h};

\node[draw=none,anchor=south, below=of n2, yshift=1cm] (x) { \footnotesize \;\;\;training time};
\node[draw=none,anchor=south, below=of n2, yshift=0.7cm] (x) { \footnotesize (GPU-h)};

\node[draw=none, circle, inner sep=6, anchor=center,fill=black] (nf) at (axis cs:\legpos,67) {};
\node[draw=none, circle, inner sep=4, anchor=center, fill=black!50] (np) at (axis cs:\legpos,67) {};
\draw [-] (np.center) -- ++ (0.8cm,0cm);
\draw [-, black] (nf.north) ++ (0cm, -0mm) -- ++ (0.8cm,0cm);
\node[draw=none,anchor=north, right=0.8cm of np.center, text=black] (x) {\footnotesize pre-training};
\node[draw=none,anchor=west, below right=-3mm and 0.8cm of nf.north] (x) {\footnotesize  fine-tuning};

\end{semilogxaxis}
\end{tikzpicture}
        \caption{{\bf Efficiency.} We report the best performance of different models between TreeSatAI and TreeSatAI-TS, with pre-training and fine-tuning using $100$\% of labels. 
        The area of the markers is proportional to the training time, broken down in pre-training and fine-tuning when applicable. }
    \label{fig:efficiency}
\end{figure}

\parag{\bf Experiments on PASTIS-HD.} 
The analysis of the performance of various models on PASTIS-HD is reported in \cref{tab:pastis}, and is consistent with the ones of TreeSatAI-TS.
First, by learning to combine all modalities despite their different resolutions,  OmniSat achieves state-of-the-art results on this benchmark. Second, our cross-modal pretraining significantly improves OmniSat's performance in the multimodal ($+10.8$ pF1-score with 100\% of training label) and all  single-modality settings ($8.8$ points for Sentinel-1, $10.7$ for Sentinel-2, and $6.5$ for the VHR images).
 
\parag{\bf Experiments on FLAIR.}
We report in \cref{tab:flair} the results on the bimodal FLAIR dataset for multilabel classification. OmniSat outperforms the much larger ScaleMAE \cite{reed2023scale} and UT\&T \cite{garioud2023flair} models with $100$\% of labels and both modalities by $3.4$\%.
Our pre-training scheme had a smaller impact than for the TreeSatAI-TS experiment. We attribute this to the fact that only two modalities are available, which decreases the supervisory power of our cross-modal contrastive objective and our multimodal reconstruction loss. This highlights a limitation of OmniSat: the model needs to be pre-trained on a modality-rich dataset to achieve its best performance. 

\parag{\bf Efficiency Evaluation.} We plot in \cref{fig:efficiency} the best performance between TreeSatAI and TreeSatAI-TS for different models according to their size and training time. OmniSat is more compact, faster to train, and performs better than all evaluated models, including the DOFA foundation model. The highly-specialized combination of PSE, LTAE, and ResNet is a strong contender, outperforming significantly larger models with generic encoding-decoding schemes.

\if 1 0
\begin{figure}
    \centering
        \begin{tikzpicture}
\begin{axis}[
    xlabel={\% annotated},
    ylabel={mIoU},
    xmin=0, xmax=100,
    ymin=20, ymax=80,
    xtick={1,5,10,25,50,100},
    ytick={20,30,40,50,60,70,80},
    legend pos=south east,
    grid style=dashed,
    ymajorgrids=true,
    xmajorgrids=true,
    grid style=dashed,
]

 \addplot[blue, ultra thick, dotted, mark=*, mark options={solid}] coordinates {(1,23.9) (5,47.2) (10,52.2) (25,63.7) (50,69.2) (100,73.3)};

 \addplot[blue, ultra thick, mark=square] coordinates {(1,36.8) (5,48.0) (10,61.1) (25,62.3) (50,69.3) (100,74.2)};


   
\legend{\textbf{OmniSAT (scratch)}, \textbf{OmniSAT (pretrained)},  SatMAE (pretrained), ScaleMAE (pretrained),}

\end{axis}
\end{tikzpicture}
    \caption{{\bf Impact of pre-training.}}
    \label{fig:enter-label}
\end{figure}
\fi

\if 1 0
\fi
\subsection{Ablation Study}
\label{sec:exp:ablation}

In this section, we report the results of several experiments evaluating the impact and validity of our main design choices, see \cref{tab:ablation}.

\parag{\bf a) Encoder/Decoder Architecture.}
We propose several improvements to the standard image encoder-decoder scheme used in computer vision to accommodate the specificities of EO data. In particular, passing the max-pool indices from the image patch encoder to its decoder allows the learned representation to focus on characterizing the spectral signature instead of fine-grained spatial information, and leads to a performance increase of $0.7$\% in the full supervision setting.

As clouds frequently obstruct optical time series, we use a unsupervised date-filtering scheme to reconstruct only meaningful acquisitions. This approach leads to a significant improvement of $3.6$\%, showcasing the benefit of developing modality-aware approaches for EO.

\parag{\bf b) Role of Loss Functions.}
When training without contrastive loss, we observe a decrease in performance of $0.8$\% in the fully supervised regime, and a more pronounced drop of $5.5$\% in the semi-supervised regime. This demonstrates how learning consistent encoding across encoders facilitates their subsequent fusion. Interestingly, when implementing a naive contrastive loss that considers all negative examples from the batch, the decrease is greater than simply removing this loss ($2$\% in full supervision). 
This strategy may introduce indistinguishable negative examples and perturb the learning process.

We also remove the reconstruction loss, meaning that only the encoders are learned contrastively during pre-training. This results in a drop of $2$\% F1-score point, illustrating the importance of pre-training the transformer $\comb$ alongside its encoders.

\parag{\bf Limitations.} 
All datasets used in our experiments are based in Europe, primarily due to the availability of open-access annotations. This regional focus prevents us from evaluating our model's performance in tropical and developing countries, which present unique challenges in terms of label provision, heterogeneity, and complex classes. 

A limitation of our pre-training scheme is its dependence on a sufficient number of aligned modalities, as illustrated by its moderate impact on the bimodal FLAIR dataset.


\begin{table}[t]   
\caption{{\bf Ablation Study.} We present the impact of several design choices on the TreeSatAI-TS dataset, measured in terms of macro-averaged F1-score.\vspace{-1mm}}
    \label{tab:ablation}
    \centering
   \begin{tabular}{lcc@{\qquad}lcc}
    \toprule
    \multicolumn{1}{l}{Experiment} & 10\% & 100\% & \multicolumn{1}{l}{Experiment}  & 10\% & 100\%
    \\\midrule
       \bf OmniSat & \bf 61.1 & \bf 74.2 & 
        b) no contrastive loss  & 55.6 & 73.4
       \\  
       a) no index bypass & 57.5 &73.5 &
       b) naive contrastive loss & 57.8 & 72.2
       \\
       a) no date filtering  & 58.2  & 71.6&
      
       b) no reconstruction loss  & 59.0 & 72.2
       \\
       \bottomrule
    \end{tabular}
\end{table}

\section{Conclusion}
We introduced OmniSat, a new architecture for the self-supervised modality fusion of Earth Observation (EO) data from multiple sources. To facilitate its evaluation, we augmented {two} existing datasets with new modalities of different natures and resolutions. We experimentally showed that leveraging diverse modalities with a flexible model improves the model's performance in both fully and semi-supervised settings. Moreover, our training scheme can exploit the spatial alignment of multiple modalities to improve our model's unimodal performance. Finally, we proposed several improvements to leverage the unique structure of EO data in the architecture of our model, such as automatic date filtering for reconstructing time series. 
We hope that our promising results and new datasets will encourage the computer vision community to consider EO data as a playing field for evaluating and developing novel self-supervised multimodal algorithms.
\pagebreak


\section*{Acknowledgements} 
This work was supported by ANR project READY3D ANR-19-CE23-0007, and was granted access to the HPC resources of IDRIS under the allocations AD011014719 and AD011014286R1 made by GENCI.
We thank Anatol Garioud and Sébastien Giordano for their help on the creation of TreeSatAI-TS and PASTIS-HD datasets.
The SPOT images are opendata thanks to the Dataterra Dinamis initiative in the case of the \href{https://dinamis.data-terra.org/opendata/}{"Couverture France DINAMIS" program}. 
We thank Jordi Inglada for inspiring discussions and valuable feedback.


\bibliographystyle{splncs04}
\bibliography{mybib}

\begin{thebibliography}{100}
\providecommand{\url}[1]{\texttt{#1}}
\providecommand{\urlprefix}{URL }
\providecommand{\doi}[1]{https://doi.org/#1}

\bibitem{reduce}
{PyTorch: ReduceLROnPlateau}. \url{org/docs/stable/generated/torch.optim.lr_scheduler.ReduceLROnPlateau.html\#torch.optim.lr_scheduler.ReduceLROnPlateau}, accessed: 2024-02-29

\bibitem{ahlswede2022treesatai}
Ahlswede, S., Schulz, C., Gava, C., Helber, P., Bischke, B., F{\"o}rster, M., Arias, F., Hees, J., Demir, B., Kleinschmit, B.: {TreeSatAI Benchmark Archive: A} multi-sensor, multi-label dataset for tree species classification in remote sensing. Earth System Science Data Discussions  (2022)

\bibitem{alayrac2022flamingo}
Alayrac, J.B., Donahue, J., Luc, P., Miech, A., Barr, I., Hasson, Y., Lenc, K., Mensch, A., Millican, K., Reynolds, M., Ring, R., Rutherford, E., Cabi, S., Han, T., Gong, Z., Samangooei, S., Monteiro, M., Menick, J., Borgeaud, S., Brock, A., Nematzadeh, A., Sharifzadeh, S., Binkowski, M., Barreira, R., Vinyals, O., Zisserman, A., Simonyan, K.: Flamingo: {A} visual language model for few-shot learning. In: NeurIPS (2022)

\bibitem{amitrano2021earth}
Amitrano, D., Di~Martino, G., Guida, R., Iervolino, P., Iodice, A., Papa, M.N., Riccio, D., Ruello, G.: Earth environmental monitoring using multi-temporal synthetic aperture radar: {A} critical review of selected applications. Remote Sensing  (2021)

\bibitem{anderson2017earth}
Anderson, K., Ryan, B., Sonntag, W., Kavvada, A., Friedl, L.: Earth observation in service of the 2030 agenda for sustainable development. Geo-spatial Information Science  (2017)

\bibitem{assran2023self}
Assran, M., Duval, Q., Misra, I., Bojanowski, P., Vincent, P., Rabbat, M., LeCun, Y., Ballas, N.: Self-supervised learning from images with a joint-embedding predictive architecture. In: CVPR (2023)

\bibitem{ayush2021geography}
Ayush, K., Uzkent, B., Meng, C., Tanmay, K., Burke, M., Lobell, D., Ermon, S.: Geography-aware self-supervised learning. In: ICCV (2021)

\bibitem{badrinarayanan2017segnet}
Badrinarayanan, V., Kendall, A., Cipolla, R.: Segnet: {A} deep convolutional encoder-decoder architecture for image segmentation. IEEE TPAMI  (2017)

\bibitem{baevski2022data2vec}
Baevski, A., Hsu, W.N., Xu, Q., Babu, A., Gu, J., Auli, M.: Data2vec: {A} general framework for self-supervised learning in speech, vision and language. In: ICML (2022)

\bibitem{bao2021beit}
Bao, H., Dong, L., Piao, S., Wei, F.: {BEiT: BERT} pre-training of image transformers. In: ICLR (2021)

\bibitem{BAO202386}
Bao, X., Zhang, R., Lv, J., Wu, R., Zhang, H., Chen, J., Zhang, B., Ouyang, X., Liu, G.: Vegetation descriptors from {S}entinel-1 {SAR} data for crop growth monitoring. ISPRS Journal of Photogrammetry and Remote Sensing  (2023)

\bibitem{bastani2023satlaspretrain}
Bastani, F., Wolters, P., Gupta, R., Ferdinando, J., Kembhavi, A.: {SatlasPretrain: A} large-scale dataset for remote sensing image understanding. In: ICCV (2023)

\bibitem{bayoudh2022survey}
Bayoudh, K., Knani, R., Hamdaoui, F., Mtibaa, A.: A survey on deep multimodal learning for computer vision: {A}dvances, trends, applications, and datasets. The Visual Computer  (2022)

\bibitem{benedetti2018m}
Benedetti, P., Ienco, D., Gaetano, R., Ose, K., Pensa, R.G., Dupuy, S.: M$^{3}$-fusion: {A} deep learning architecture for multiscale multimodal multitemporal satellite data fusion. IEEE Journal of Selected Topics in Applied Earth Observations and Remote Sensing  (2018)

\bibitem{caron2020unsupervised}
Caron, M., Misra, I., Mairal, J., Goyal, P., Bojanowski, P., Joulin, A.: Unsupervised learning of visual features by contrasting cluster assignments. In: NeurIPS (2020)

\bibitem{caron2021emerging}
Caron, M., Touvron, H., Misra, I., J{\'e}gou, H., Mairal, J., Bojanowski, P., Joulin, A.: Emerging properties in self-supervised vision transformers. In: ICCV (2021)

\bibitem{chen2020simple}
Chen, T., Kornblith, S., Norouzi, M., Hinton, G.: A simple framework for contrastive learning of visual representations. In: ICML (2020)

\bibitem{chen2021exploring}
Chen, X., He, K.: Exploring simple siamese representation learning. In: CVPR (2021)

\bibitem{christie2018functional}
Christie, G., Fendley, N., Wilson, J., Mukherjee, R.: Functional map of the world. In: CVPR (2018)

\bibitem{cong2022satmae}
Cong, Y., Khanna, S., Meng, C., Liu, P., Rozi, E., He, Y., Burke, M., Lobell, D., Ermon, S.: {SatMAE: P}re-training transformers for temporal and multi-spectral satellite imagery. In: NeurIPS (2022)

\bibitem{coppin2002digital}
Coppin, P., Lambin, E., Jonckheere, I., Muys, B.: Digital change detection methods in natural ecosystem monitoring:{ A} review. Analysis of multi-temporal remote sensing images  (2002)

\bibitem{corley2023revisiting}
Corley, I., Robinson, C., Dodhia, R., Ferres, J.M.L., Najafirad, P.: Revisiting pre-trained remote sensing model benchmarks: {R}esizing and normalization matters. arXiv preprint arXiv:2305.13456  (2023)

\bibitem{dai20183dmv}
Dai, A., Nie{\ss}ner, M.: {3DMV}: {J}oint {3D}-multi-view prediction for {3D} semantic scene segmentation. In: ECCV (2018)

\bibitem{spot}
{DataTerra Dinamis}: Diffusion {OpenData} {Dinamis}, \url{https://dinamis.data-terra.org/opendata/}, accessed: 2023-12-15

\bibitem{dosovitskiy2020image}
Dosovitskiy, A., Beyer, L., Kolesnikov, A., Weissenborn, D., Zhai, X., Unterthiner, T., Dehghani, M., Minderer, M., Heigold, G., Gelly, S., et~al.: An image is worth 16x16 words: {T}ransformers for image recognition at scale. In: ICLR (2020)

\bibitem{Sentinel-2-paper}
Drusch, M., {Del Bello}, U., Carlier, S., Colin, O., Fernandez, V., Gascon, F., Hoersch, B., Isola, C., Laberinti, P., Martimort, P., Meygret, A., Spoto, F., Sy, O., Marchese, F., Bargellini, P.: Sentinel-2: {ESA}'s optical high-resolution mission for {GMES} operational services. Remote Sensing of Environment  (2012)

\bibitem{ebel2022sen12ms}
Ebel, P., Xu, Y., Schmitt, M., Zhu, X.X.: {SEN12MS-CR-TS: A} remote-sensing data set for multimodal multitemporal cloud removal. IEEE TGRS  (2022)

\bibitem{ekim2023mapinwild}
Ekim, B., Stomberg, T.T., Roscher, R., Schmitt, M.: {MapInWild: A} remote sensing dataset to address the question of what makes nature wild. IEEE Geoscience and Remote Sensing Magazine  (2023)

\bibitem{fuller2023croma}
Fuller, A., Millard, K., Green, J.R.: {CROMA: R}emote sensing representations with contrastive radar-optical masked autoencoders. In: NeurIPS (2023)

\bibitem{gao2022general}
Gao, Y., Sun, X., Liu, C.: A general self-supervised framework for remote sensing image classification. Remote Sensing  (2022)

\bibitem{garioud2023flair}
Garioud, A., Gonthier, N., Landrieu, L., De~Wit, A., Valette, M., Poup{\'e}e, M., Giordano, S., Wattrelos, B.: {FLAIR: A} country-scale land cover semantic segmentation dataset from multi-source optical imagery. In: NeurIPS Dataset and Benchmark (2023)

\bibitem{garnot2020lightweight}
Garnot, V.S.F., Landrieu, L.: Lightweight temporal self-attention for classifying satellite images time series. In: Advanced Analytics and Learning on Temporal Data: ECML PKDD Workshop (2020)

\bibitem{garnot2021panoptic}
Garnot, V.S.F., Landrieu, L.: Panoptic segmentation of satellite image time series with convolutional temporal attention networks. In: ICCV (2021)

\bibitem{garnot2022multi}
Garnot, V.S.F., Landrieu, L., Chehata, N.: Multi-modal temporal attention models for crop mapping from satellite time series. ISPRS Journal of Photogrammetry and Remote Sensing  (2022)

\bibitem{garnot2020satellite}
Garnot, V.S.F., Landrieu, L., Giordano, S., Chehata, N.: Satellite image time series classification with pixel-set encoders and temporal self-attention. In: CVPR (2020)

\bibitem{ghamisi2019multisource}
Ghamisi, P., Rasti, B., Yokoya, N., Wang, Q., Hofle, B., Bruzzone, L., Bovolo, F., Chi, M., Anders, K., Gloaguen, R., et~al.: Multisource and multitemporal data fusion in remote sensing: {A} comprehensive review of the state of the art. IEEE Geoscience and Remote Sensing Magazine  (2019)

\bibitem{gidaris2018unsupervised}
Gidaris, S., Singh, P., Komodakis, N.: Unsupervised representation learning by predicting image rotations. In: ICLR (2018)

\bibitem{girdhar2023imagebind}
Girdhar, R., El-Nouby, A., Liu, Z., Singh, M., Alwala, K.V., Joulin, A., Misra, I.: Imagebind: {O}ne embedding space to bind them all. In: CVPR (2023)

\bibitem{girdhar2022omnivore}
Girdhar, R., Singh, M., Ravi, N., van~der Maaten, L., Joulin, A., Misra, I.: Omnivore:{ A} single model for many visual modalities. In: CVPR (2022)

\bibitem{goldberg2023automated}
Goldberg, H.R., Ratto, C.R., Banerjee, A., Kelbaugh, M.T., Giglio, M., Vermote, E.F.: Automated global-scale detection and characterization of anthropogenic activity using multi-source satellite-based remote sensing imagery. In: Geospatial Informatics XIII. SPIE (2023)

\bibitem{greenwell2024watch}
Greenwell, C., Crall, J., Purri, M., Dana, K., Jacobs, N., Hadzic, A., Workman, S., Leotta, M.: {WATCH: W}ide-area terrestrial change hypercube. In: WACV (2024)

\bibitem{grill2020bootstrap}
Grill, J.B., Strub, F., Altch{\'e}, F., Tallec, C., Richemond, P., Buchatskaya, E., Doersch, C., Avila~Pires, B., Guo, Z., Gheshlaghi~Azar, M., et~al.: Bootstrap your own latent-a new approach to self-supervised learning. In: NeurIPS (2020)

\bibitem{hackstein2024exploring}
Hackstein, J., Sumbul, G., Clasen, K.N., Demir, B.: Exploring masked autoencoders for sensor-agnostic image retrieval in remote sensing. arXiv preprint arXiv:2401.07782  (2024)

\bibitem{hazirbas2017fusenet}
Hazirbas, C., Ma, L., Domokos, C., Cremers, D.: Fuse{N}et: {I}ncorporating depth into semantic segmentation via fusion-based {CNN} architecture. In: ACCV (2017)

\bibitem{he2022masked}
He, K., Chen, X., Xie, S., Li, Y., Doll{\'a}r, P., Girshick, R.: Masked autoencoders are scalable vision learners. In: CVPR (2022)

\bibitem{he2020momentum}
He, K., Fan, H., Wu, Y., Xie, S., Girshick, R.: Momentum contrast for unsupervised visual representation learning. In: CVPR (2020)

\bibitem{he2016deep}
He, K., Zhang, X., Ren, S., Sun, J.: Deep residual learning for image recognition. In: CVPR (2016)

\bibitem{hong2020more}
Hong, D., Gao, L., Yokoya, N., Yao, J., Chanussot, J., Du, Q., Zhang, B.: More diverse means better: {M}ultimodal deep learning meets remote-sensing imagery classification. IEEE TGRS  (2020)

\bibitem{hu2022mdas}
Hu, J., Liu, R., Hong, D., Camero, A., Yao, J., Schneider, M., Kurz, F., Segl, K., Zhu, X.X.: {MDAS: A }new multimodal benchmark dataset for remote sensing. Earth System Science Data Discussions  (2022)

\bibitem{huang2022mavil}
Huang, P.Y., Sharma, V., Xu, H., Ryali, C., Fan, H., Li, Y., Li, S.W., Ghosh, G., Malik, J., Feichtenhofer, C.: {MAViL: M}asked audio-video learners. In: NeurIPS (2023)

\bibitem{ibanez2022masked}
Ibanez, D., Fernandez-Beltran, R., Pla, F., Yokoya, N.: Masked auto-encoding spectral--spatial transformer for hyperspectral image classification. IEEE TGRS  (2022)

\bibitem{irvin2023usat}
Irvin, J., Tao, L., Zhou, J., Ma, Y., Nashold, L., Liu, B., Ng, A.Y.: {USat: A} unified self-supervised encoder for multi-sensor satellite imagery. arXiv preprint arXiv:2312.02199  (2023)

\bibitem{kenton2019bert}
Kenton, J.D.M.W.C., Toutanova, L.K.: {BERT: P}re-training of deep bidirectional transformers for language understanding. In: NAACL (2019)

\bibitem{kingma2014adam}
Kingma, D.P., Ba, J.: {ADAM: A} method for stochastic optimization. ICLR  (2015)

\bibitem{krispel2020fuseseg}
Krispel, G., Opitz, M., Waltner, G., Possegger, H., Bischof, H.: {Fuseseg: LiDAR} point cloud segmentation fusing multi-modal data. In: WACV (2020)

\bibitem{kuffer2020role}
Kuffer, M., Thomson, D.R., Boo, G., Mahabir, R., Grippa, T., Vanhuysse, S., Engstrom, R., Ndugwa, R., Makau, J., Darin, E., et~al.: The role of {E}arth observation in an integrated deprived area mapping “system” for low-to-middle income countries. Remote sensing  (2020)

\bibitem{kussul2017deep}
Kussul, N., Lavreniuk, M., Skakun, S., Shelestov, A.: Deep learning classification of land cover and crop types using remote sensing data. IEEE Geoscience and Remote Sensing Letters  (2017)

\bibitem{lacoste2021toward}
Lacoste, A., Sherwin, E.D., Kerner, H., Alemohammad, H., L{\"u}tjens, B., Irvin, J., Dao, D., Chang, A., Gunturkun, M., Drouin, A., et~al.: Toward foundation models for {E}arth monitoring: {P}roposal for a climate change benchmark. arXiv preprint arXiv:2112.00570  (2021)

\bibitem{li2012current}
Li, D., Tong, Q., Li, R., Gong, J., Zhang, L.: Current issues in high-resolution {E}arth observation technology. Science China Earth sciences  (2012)

\bibitem{li2022deep}
Li, J., Hong, D., Gao, L., Yao, J., Zheng, K., Zhang, B., Chanussot, J.: Deep learning in multimodal remote sensing data fusion: {A} comprehensive review. International Journal of Applied Earth Observation and Geoinformation  \textbf{112},  102926 (2022)

\bibitem{liao2022kitti}
Liao, Y., Xie, J., Geiger, A.: {KITTI-360: A }novel dataset and benchmarks for urban scene understanding in {2D} and {3D}. IEEE TPAMI  (2022)

\bibitem{liu2022band}
Liu, Y., Li, X., Hua, Z., Xia, C., Zhao, L.: A band selection method with masked convolutional autoencoder for hyperspectral image. IEEE Geoscience and Remote Sensing Letters  (2022)

\bibitem{ma2021outcome}
Ma, Y., Li, Y., Feng, K., Xia, Y., Huang, Q., Zhang, H., Prieur, C., Licciardi, G., Malha, H., Chanussot, J., et~al.: {The outcome of the 2021 IEEE GRSS data fusion contest-Track DSE: D}etection of settlements without electricity. IEEE Journal of Selected Topics in Applied Earth Observations and Remote Sensing  (2021)

\bibitem{mai2023opportunities}
Mai, G., Huang, W., Sun, J., Song, S., Mishra, D., Liu, N., Gao, S., Liu, T., Cong, G., Hu, Y., et~al.: On the opportunities and challenges of foundation models for geospatial artificial intelligence. arXiv preprint arXiv:2304.06798  (2023)

\bibitem{manas2021seasonal}
Manas, O., Lacoste, A., Gir{\'o}-i Nieto, X., Vazquez, D., Rodriguez, P.: Seasonal contrast: {U}nsupervised pre-training from uncurated remote sensing data. In: ICCV (2021)

\bibitem{manfreda2018use}
Manfreda, S., McCabe, M.F., Miller, P.E., Lucas, R., Pajuelo~Madrigal, V., Mallinis, G., Ben~Dor, E., Helman, D., Estes, L., Ciraolo, G., et~al.: On the use of unmanned aerial systems for environmental monitoring. Remote sensing p.~641 (2018)

\bibitem{moreira2013tutorial}
Moreira, A., Prats-Iraola, P., Younis, M., Krieger, G., Hajnsek, I., Papathanassiou, K.P.: A tutorial on synthetic aperture radar. IEEE Geoscience and Remote Sensing Magazine  (2013)

\bibitem{nakalembe2020urgent}
Nakalembe, C.: Urgent and critical need for sub-saharan african countries to invest in {E}arth observation-based agricultural early warning and monitoring systems. Environmental Research Letters  (2020)

\bibitem{Silberman:ECCV12}
Nathan~Silberman, Derek~Hoiem, P.K., Fergus, R.: Indoor segmentation and support inference from {RGBD} images. In: ECCV (2012)

\bibitem{noroozi2016unsupervised}
Noroozi, M., Favaro, P.: Unsupervised learning of visual representations by solving jigsaw puzzles. In: ECCV (2016)

\bibitem{oord2018representation}
Oord, A.v.d., Li, Y., Vinyals, O.: Representation learning with contrastive predictive coding. arXiv preprint arXiv:1807.03748  (2018)

\bibitem{oquab2023dinov2}
Oquab, M., Darcet, T., Moutakanni, T., Vo, H., Szafraniec, M., Khalidov, V., Fernandez, P., Haziza, D., Massa, F., El-Nouby, A., et~al.: Dinov2: {L}earning robust visual features without supervision. TLMR  (2023)

\bibitem{pohl1998review}
Pohl, C., Van~Genderen, J.L.: Multisensor image fusion in remote sensing: {C}oncepts, methods and applications. International journal of remote sensing  (1998)

\bibitem{radford2021learning}
Radford, A., Kim, J.W., Hallacy, C., Ramesh, A., Goh, G., Agarwal, S., Sastry, G., Askell, A., Mishkin, P., Clark, J., et~al.: Learning transferable visual models from natural language supervision. In: ICML (2021)

\bibitem{recasens2023zorro}
Recasens, A., Lin, J., Carreira, J., Jaegle, D., Wang, L., Alayrac, J.b., Luc, P., Miech, A., Smaira, L., Hemsley, R., et~al.: Zorro: {T}he masked multimodal transformer. arXiv preprint arXiv:2301.09595  (2023)

\bibitem{reed2023scale}
Reed, C.J., Gupta, R., Li, S., Brockman, S., Funk, C., Clipp, B., Keutzer, K., Candido, S., Uyttendaele, M., Darrell, T.: {Scale-MAE: A} scale-aware masked autoencoder for multiscale geospatial representation learning. In: ICCV (2023)

\bibitem{robert2022learning}
Robert, D., Vallet, B., Landrieu, L.: Learning multi-view aggregation in the wild for large-scale 3{D} semantic segmentation. In: CVPR (2022)

\bibitem{robinson2021global}
Robinson, C., Malkin, K., Jojic, N., Chen, H., Qin, R., Xiao, C., Schmitt, M., Ghamisi, P., H{\"a}nsch, R., Yokoya, N.: Global land-cover mapping with weak supervision: {O}utcome of the 2020 {IEEE GRSS} data fusion contest. IEEE Journal of Selected Topics in Applied Earth Observations and Remote Sensing  (2021)

\bibitem{rolf2021generalizable}
Rolf, E., Proctor, J., Carleton, T., Bolliger, I., Shankar, V., Ishihara, M., Recht, B., Hsiang, S.: A generalizable and accessible approach to machine learning with global satellite imagery. Nature communications  (2021)

\bibitem{russwurm2020self}
Ru{\ss}wurm, M., K{\"o}rner, M.: Self-attention for raw optical satellite time series classification. ISPRS Journal of Photogrammetry and Remote Sensing  (2020)

\bibitem{schmitt2016data}
Schmitt, M., Zhu, X.X.: Data fusion and remote sensing: {A}n ever-growing relationship. IEEE Geoscience and Remote Sensing Magazine  (2016)

\bibitem{secades2014earth}
Secades, C., O'Connor, B., Brown, C., Walpole, M., et~al.: Earth observation for biodiversity monitoring: {A} review of current approaches and future opportunities for tracking progress towards the aichi biodiversity targets. CBD technical series  (2014)

\bibitem{shermeyer2020spacenet}
Shermeyer, J., Hogan, D., Brown, J., Van~Etten, A., Weir, N., Pacifici, F., Hansch, R., Bastidas, A., Soenen, S., Bacastow, T., et~al.: Spacenet 6: {M}ulti-sensor all weather mapping dataset. In: CVPR Workshop EarthVision (2020)

\bibitem{shukor2023unival}
Shukor, M., Dancette, C., Rame, A., Cord, M.: {UnIVAL: U}nified model for image, video, audio and language tasks. TMLR  (2023)

\bibitem{skidmore2021priority}
Skidmore, A.K., Coops, N.C., Neinavaz, E., Ali, A., Schaepman, M.E., Paganini, M., Kissling, W.D., Vihervaara, P., Darvishzadeh, R., Feilhauer, H., et~al.: Priority list of biodiversity metrics to observe from space. Nature Ecology \& Evolution  (2021)

\bibitem{srivastava2024omnivec}
Srivastava, S., Sharma, G.: {OmniVec}: {L}earning robust representations with cross modal sharing. In: WACV (2024)

\bibitem{sudmanns2019assessing}
Sudmanns, M., Tiede, D., Augustin, H., Lang, S.: Assessing global {S}entinel-2 coverage dynamics and data availability for operational {E}arth observation ({EO}) applications using the {EO-Compass}. International Journal of Digital {E}arth  (2019)

\bibitem{sumbul2021bigearthnet}
Sumbul, G., De~Wall, A., Kreuziger, T., Marcelino, F., Costa, H., Benevides, P., Caetano, M., Demir, B., Markl, V.: {BigEarthNet-MM: A} large-scale, multimodal, multilabel benchmark archive for remote sensing image classification and retrieval. IEEE Geoscience and Remote Sensing Magazine  (2021)

\bibitem{tarasiou2023vits}
Tarasiou, M., Chavez, E., Zafeiriou, S.: {ViTs for SITS: V}ision transformers for satellite image time series. In: CVPR (2023)

\bibitem{tseng2023lightweight}
Tseng, G., Zvonkov, I., Purohit, M., Rolnick, D., Kerner, H.: Lightweight, pre-trained transformers for remote sensing timeseries. arXiv preprint arXiv:2304.14065  (2023)

\bibitem{tseng2022croco}
Tseng, W.H., L{\^e}, H.{\^A}., Boulch, A., Lef{\`e}vre, S., Tiede, D.: {CROCO}: {C}ross-modal contrastive learning for localization of {E}arth observation data. ISPRS Annals of the Photogrammetry, Remote Sensing and Spatial Information Sciences  (2022)

\bibitem{vaswani2017attention}
Vaswani, A., Shazeer, N., Parmar, N., Uszkoreit, J., Jones, L., Gomez, A.N., Kaiser, {\L}., Polosukhin, I.: Attention is all you need. In: NeurIPS (2017)

\bibitem{vrieling2018vegetation}
Vrieling, A., Meroni, M., Darvishzadeh, R., Skidmore, A.K., Wang, T., Zurita-Milla, R., Oosterbeek, K., O'Connor, B., Paganini, M.: Vegetation phenology from {S}entinel-2 and field cameras for a {D}utch barrier island. Remote sensing of environment  (2018)

\bibitem{wang2022ssl4eo}
Wang, Y., Braham, N.A.A., Xiong, Z., Liu, C., Albrecht, C.M., Zhu, X.X.: {SSL4EO-S12}: {A} large-scale multi-modal, multi-temporal dataset for self-supervised learning in {E}arth observation. IEEE Geoscience and Remote Sensing Magazine  (2023)

\bibitem{wenger2022multisenge}
Wenger, R., Puissant, A., Weber, J., Idoumghar, L., Forestier, G.: {MultiSen{GE}: A} multimodal and multitemporal benchmark dataset for land use/land cover remote sensing applications. ISPRS Annals of the Photogrammetry, Remote Sensing and Spatial Information Sciences  (2022)

\bibitem{wu2021rethinking}
Wu, K., Peng, H., Chen, M., Fu, J., Chao, H.: Rethinking and improving relative position encoding for vision transformer. In: ICCV (2021)

\bibitem{xie2022simmim}
Xie, Z., Zhang, Z., Cao, Y., Lin, Y., Bao, J., Yao, Z., Dai, Q., Hu, H.: {SimMim: A} simple framework for masked image modeling. In: CVPR (2022)

\bibitem{xiong2024dofa}
Xiong, Z., Wang, Y., Zhang, F., Stewart, A.J., Hanna, J., Borth, D., Papoutsis, I., Saux, B.L., Camps-Valls, G., Zhu, X.X.: Neural plasticity-inspired foundation model for observing the {E}arth crossing modalities. arXiv preprint arXiv:2403.15356  (2024)

\bibitem{yang2013role}
Yang, J., Gong, P., Fu, R., Zhang, M., Chen, J., Liang, S., Xu, B., Shi, J., Dickinson, R.: The role of satellite remote sensing in climate change studies. Nature climate change  (2013)

\bibitem{yang2021muti}
Yang, M.Y., Landrieu, L., Tuia, D., Toth, C.: Muti-modal learning in photogrammetry and remote sensing. ISPRS Journal of Photogrammetry and Remote Sensing  (2021)

\bibitem{yuan2022sits}
Yuan, Y., Lin, L., Liu, Q., Hang, R., Zhou, Z.G.: {SITS-Former: A} pre-trained spatio-spectral-temporal representation model for sentinel-2 time series classification. International Journal of Applied Earth Observation and Geoinformation  (2022)

\bibitem{zhang2016colorful}
Zhang, R., Isola, P., Efros, A.A.: Colorful image colorization. In: ECCV (2016)

\bibitem{zhou2022image}
Zhou, J., Wei, C., Wang, H., Shen, W., Xie, C., Yuille, A., Kong, T.: Image {BERT} pre-training with online tokenizer. In: ICLR (2022)

\bibitem{zong2023self}
Zong, Y., Mac~Aodha, O., Hospedales, T.: Self-supervised multimodal learning: {A} survey. arXiv preprint arXiv:2304.01008  (2023)

\end{thebibliography}

 \ARXIV{
     %
     \section*{\centering \LARGE Appendix}
     \setcounter{section}{0}
     \setcounter{figure}{0}
     \setcounter{table}{0}
     \renewcommand*{\theHsection}{appendix.\the\value{section}}
     \renewcommand\thefigure{\arabic{figure}}
     \renewcommand\thetable{\arabic{table}}

\renewcommand\thefigure{A-\arabic{figure}}
\renewcommand\thesection{A-\arabic{section}}
\renewcommand\thetable{A-\arabic{table}}
\renewcommand\theequation{A-\arabic{equation}}
\setcounter{equation}{0}
\setcounter{section}{0}
\setcounter{figure}{0}
\setcounter{table}{0}

In this appendix, we present an extended ablation study in \secref{sec:supablation}, details our competing methods in \secref{sec:competing}, provide the classwise performance in \secref{sec:suresults}, and provide qualitative illustrations in \figref{fig:quali}.

\section{Supplementary Ablations}
\label{sec:supablation}
We propose supplementary ablations to evaluate  the impact of several design choices.

\begin{table}[]
\caption{{\bf Supplementary Ablation.} Performance (weighted F1) on TreeSatAI-TS of alternate VHR encoders (\textbf{a}-\textbf{c}) and masking schemes (\textbf{d}-\textbf{e}).}
    \centering
    \begin{tabular}{llcccc}
    \toprule
        \multicolumn{2}{l}{Experiment} &  \quad All\quad~  &  \quad VHR \quad~ &   \quad S2\quad~  &  \quad S1\quad~  \\
    \midrule
         \multicolumn{2}{l}{Default} & \bf 74.2 & 70.5 & 62.9 & \bf 56.7  \\ \greyrule
        \bf a &linear w. random init & 66.8 & 57.3 & 58.9 & 54.8 \\
        \bf b & ViT & 70.5 & \bf 70.8 & 64.0 & 52.6\\              
        \bf c & linear from ScaleMAE & 68.9 & 51.2 & \bf 66.7 & 52.2 \\\greyrule
         \bf d &Spatial masking & 73.2 & 70.1 & 63.2 & 54.6 \\
        \bf e & Modality masking & 72.4 & 70.2 & 61.2 & 55.4\\
    \bottomrule
    \end{tabular}
    \label{tab:xp}
    \vspace{-5.8mm}
\end{table}

\paragraph{\bf  Alternate VHR Encoder.} To train OmniSat on both VHR (0.2 m) and Sentinel (10m) images, we must embed patches of $50\!\times\!50$  pixels.
    We consider here alternative encoders to CNNs: a linear layer (\cref{tab:xp}.\textbf{a}) and a ViT with $10\!\times\!10$ patches (\cref{tab:xp}.\textbf{b}). The results suggest that $50\times50$ patches are too large to use linear projection. While ViTs reach slightly higher unimodal performance, CNNs allow us to bypass maxpool indices to the decoders (\cref{sec:archi}) leading to higher multimodal performances.
    
    \paragraph{\bf  Using Pre-trained VHR models.} Rescaling the $50\times50$ patches to the $224 \times 224$ resolution of ScaleMAE or SatMAE proved impractical in terms of memory. Instead, we use the pre-trained patch encoder of ScaleMAE by rescaling our $50\!\times\!50$ patches to $16\!\times\!16$, removing the infrared channel, and adding a projection layer to our token size $D=256$ (\cref{tab:xp}.\textbf{c}). Interestingly, this leads to a cross-modal distillation which improves the results for S2. The VHR and multimodal performance remain below OmniSat, which can be attributed to the lack of a NIR channel.
    
    \paragraph{\bf  Masking Strategies.} We report the results for spatially consistent masking (patches are masked for all modalities simultaneously, \cref{tab:xp}.\textbf{d}) and modality masking (the patches of a random modality are all masked, \cref{tab:xp}.\textbf{e}). Our random masking strategy performs better.

    \paragraph{\bf Relative \vs Absolute Positional Encoding.}
    We evaluate the impact of replacing the relative positional encoding of tokens, based on the patch position, with an absolute position encoding, based on the position of the patches in their tile---similar to what is classically done for image processing.

    With an absolute positional encoding, OmniSat reaches an F1-score of $58.4$ and $73.0$ when fine-tuned with $10$\% and $100$\% of the training set of TreeSatAI-TS, respectively. This is $2.7$ and $1.2$\% below a model trained with relative positional encodings. We conclude that relative positional encodings are better suited for analyzing EO images. While the upper patches of natural images are bound to correspond to the sky, and the lower patches contain ground, no such analogy can be made for EO data, whose distribution is equivariant through small horizontal translation. 
    
    
    \paragraph{\bf Impact of Pre-training on Monomodal Performance.}
    We aim to determine how our multimodal pre-training scheme improves the monomodal performance (\eg, +13.2\% for Sentinel-2 in full supervision). We consider two mechanisms that may lead to more discriminative features: (i) multimodality allows us to train the modality combiner network $\comb$ with more data, or 
    (ii) our cross-modal and token-wise alignment-based losses provide a strong supervisory signal. We propose an experiment to verify which mechanism is the leading reason of our scheme's strong performance. 

    We pre-train OmniSat on TreeSatAI-TS in mono- and multimodal settings \emph{with a constant amount of tokens}. More precisely, we pre-train OmniSat using \emph{all} input tokens from the S2 modality \emph{only}, and using \emph{all} $3$ modalities but only $33\%$ of patches. This means that each experiment considering the same number $P$ of input tokens. We then train a single linear layer to map these representations to class scores (linear probing) using $10$ and $100$\% of the annotated S2 data.
    Finally, we evaluate the quality of these linear mappings on the test set using only the S2 modality.

    The model trained with a multimodal pretext task reaches a F1-score of $44.7$ for $10$\% and $46.3$ for $100$\% of the training data. The model trained only with S2 performs significantly worse: $26.9$ for $10$\% and $29.8$ for $100$\% of data. This result suggests that the key to the efficacy of our pretraining scheme is the supervisory signal of per-patch contrastive and reconstruction objectives, rather than just increasing the number of tokens viewed by the transformer backbone.


\section{Adapting Competing Methods}
\label{sec:competing}
We adapt competing methods to allow them to handle single images and time series at different resolutions. We performed multiple tests for each approach and kept the configurations leading to the competing approach' highest performance.
\begin{itemize}
    \item {\bf Multimodality.} We train methods that are not natively multimodal (PSE \cite{garnot2020satellite}, ViT \cite{dosovitskiy2020image}, DOFA \cite{xiong2024dofa}, SatMAE, ScaleMAE) using a late-fusion scheme \cite{hong2020more} by concatenating the embeddings learned in each modality, as suggested by Ahlswede \etal \cite{ahlswede2022treesatai}. For UT\&T \cite{garioud2023flair}, initially designed for VHR images and Sentinel-2 time series, we add a branch for Sentinel-1 integration, which is identical to the Sentinel-2 branch except for the first layer.
    \item {\bf Handling Temporal Data.} To evaluate image models  (SatMAE, ScaleMAE, CROMA) on time series, we convert image sequences to single images by concatenating for each pixel and channel 
 channel-wise the median observation for the four seasons: spring, summer, fall, and winter \cite{kussul2017deep}. 
    \item {\bf Handling VHR Data.} 
    To evaluate methods designed for low-resolution images (PSE, LTAE \cite{garnot2020lightweight}) in a multimodal setting that includes VHR images, we concatenate their final embedding to the the one of a ResNet network.
    \item {\bf Scaling Models.} The considered datasets are smaller than the ones typically used to train large ViT-based models, making them prone to overfitting. 
    To address this issue we select a ViT-Small \cite{dosovitskiy2020image} backbone for SatMAE, ScaleMAE and CROMA. {For DOFA, we use a ViT-Base, the smallest pretrained model available.}
    \item {\bf Multi-Class Prediction.} To evaluate ViT-based models on classification experiments, we insert a linear layer that maps the embedding of the class token $\langle\texttt{CLS}\rangle$ to a vector of class scores. For the UT\&T model, we compute a spatial average of the last feature map, followed by a similar linear projection.
\end{itemize}

\section{Supplementary Results}
\label{sec:suresults}

We report the performance of different approaches for each class for the two datasets graphically in \figref{fig:classwise} and as a table in \tabref{tab:classwise}. OmniSat is able to parse complex scenes including mixed forest, cultures, and complex urban areas. In particular, Omnisat leverage temporal dynamics to distinguish between different vegetation species. 

\begin{figure}
    \centering
    \begin{tikzpicture}
\begin{axis}[
    width=.9\linewidth,
    height=.25\textheight,
    ybar,
    bar width=.2cm,
    ylabel={Proportion}, 
    symbolic x coords={Pinus,Quercus,Fagus,Picea,Cleared,Larix,Pseudotsuga,Betula,Acer,Alnus,Fraxinus,Abies,Populus,Prunus,Tilia},
    xtick=data,
    nodes near coords align={vertical},
    axis y line*=left, 
    xticklabel style={rotate=50,anchor=east,font=\small},
    ymin=0,
    ymax=25,
    y label style={at={(axis description cs:+0.05,.5)},rotate=0,anchor=south},
    ]
\addplot [fill=gray] coordinates {(Pinus,23) (Quercus,22.8) (Fagus,22.6) (Picea,21.8) (Cleared,11.2) (Larix,9.7) (Pseudotsuga,8.9) (Betula,7.1) (Acer,6.5) (Alnus,6.5) (Fraxinus,5.7) (Abies,2.5) (Populus,1.8) (Prunus,0.4) (Tilia, 0.5)};
\legend{}; 
\end{axis}

\begin{axis}[
    width=.9\linewidth,
    height=.25\textheight,
    axis y line*=right, 
    axis x line=none,
    ymin=0,
    ymax=100,
    symbolic x coords={Pinus,Quercus,Fagus,Picea,Cleared,Larix,Pseudotsuga,Betula,Acer,Alnus,Fraxinus,Abies,Populus,Prunus,Tilia},
    legend style={at={(0.5,-0.50)},
      anchor=north,legend columns=-1},
    ylabel={IoU}, 
    xticklabel style={rotate=50,small,anchor=east},
    ylabel near ticks, yticklabel pos=right,
]
\addlegendimage{/pgfplots/refstyle=Proportion}\addlegendentry{Proportion}
\addplot[mark=*,very thick, UTTCOLOR] coordinates {(Pinus,72.1) (Quercus,62.5) (Fagus,60.7) (Picea,63.6) (Cleared,53.7) (Larix,43.7) (Pseudotsuga,45.4) (Betula,38.5) (Acer,32.4) (Alnus,49.8) (Fraxinus,45.9) (Abies,52.8) (Populus,44.8) (Prunus,32.4) (Tilia, 25.0)};
\addplot[mark=square*,SCALECOLOR, very thick] coordinates {(Pinus,78.2) (Quercus,64.0) (Fagus,63.8) (Picea,71.5) (Cleared,64.4) (Larix,51.7) (Pseudotsuga,55.1) (Betula,36.8) (Acer,43.4) (Alnus,42.7) (Fraxinus,38.3) (Abies,53.2) (Populus,25.4) (Prunus,11.8) (Tilia, 18.2)};
\addplot[mark=triangle*,OMNICOLOR, very thick] coordinates {(Pinus,85.4) (Quercus,76.1) (Fagus,72.9) (Picea,77.6) (Cleared,74.9) (Larix,69.5) (Pseudotsuga,68.5) (Betula,56.5) (Acer,58.9) (Alnus,62.5) (Fraxinus,60.0) (Abies,74.1) (Populus,64.6) (Prunus,46.2) (Tilia, 43.4)};
\legend{}; 
\end{axis}

\end{tikzpicture}
    \begin{tikzpicture}

\begin{axis}[
    width=.9\linewidth,
    height=.25\textheight,
    ybar,
    bar width=.2cm,
    legend style={at={(0.5,-0.20)},
      anchor=north,legend columns=-1},
    ylabel={Proportion}, 
    symbolic x coords={ deciduous,herbaceous,brushwood,perv. surface,agri. land,imperv. surface,coniferous,building,water,other,bare soil,plowed land,vineyard},
    xtick=data,
    nodes near coords align={vertical},
    axis y line*=left, 
    xticklabel style={rotate=50,anchor=east,font=\small},
    ymin=0,
    ymax=100,
    y label style={at={(axis description cs:+0.05,.5)},rotate=0,anchor=south},
    ]

\addplot [fill=gray] coordinates {(deciduous,84.2) (herbaceous,74.1) (brushwood,49.6) (perv. surface,45.0) (agri. land,39.4) (imperv. surface,37.7) (coniferous,36.0) (building,27.6) (water,12.5) (other,11.0) (bare soil,8.0) (plowed land,7.1) (vineyard,6.5)};
\label{Proportion}
\end{axis}

\begin{axis}[
    width=.9\linewidth,
    height=.25\textheight,
    axis y line*=right, 
    axis x line=none,
    ymin=0,
    symbolic x coords={deciduous,herbaceous,brushwood,perv. surface,agri. land,imperv. surface,coniferous,building,water,other,bare soil,plowed land,vineyard},
    legend style={at={(0.5,-0.55)},
      anchor=north,legend columns=-1},
    ylabel={IoU}, 
    xticklabel style={rotate=90,anchor=east},
    ylabel near ticks, yticklabel pos=right,
]
\addplot[mark=*,very thick, UTTCOLOR] coordinates {(deciduous,91.2) (herbaceous,84.7) (brushwood,66.3) (perv. surface,63.3) (agri. land,71.7) (imperv. surface,60.4) (coniferous,64.4) (building,52.4) (water,46.9) (other,26.2) (bare soil,32.7) (plowed land,27.1) (vineyard,57.1)};
\addplot[mark=square*,SCALECOLOR, very thick] coordinates {(deciduous,93.8) (herbaceous,82.9) (brushwood,66.3) (perv. surface,71.4) (agri. land,66.9) (imperv. surface,83.3) (coniferous,55.0) (building,86.0) (water,76.7) (other,43.1) (bare soil,31.5) (plowed land,32.8) (vineyard,30.6)};
\addplot[mark=triangle*,OMNICOLOR, very thick] coordinates {(deciduous,95.2) (herbaceous,87.5) (brushwood,73.4) (perv. surface,73.2) (agri. land,71.1) (imperv. surface,88.2) (coniferous,70.6) (building,91.8) (water,81.3) (other,48.5) (bare soil,54.4) (plowed land,36.0) (vineyard,72.1)};
\legend{}; 
\end{axis}

\end{tikzpicture}
    \begin{tikzpicture}

\begin{axis}[
    width=.9\linewidth,
    height=.25\textheight,
    ybar,
    bar width=.2cm,
    ylabel={Proportion}, 
    symbolic x coords={Meadow,Winter wheat,Corn,Winter barley,Leguminous, Winter rapeseed,Fruits\, veg.\, flow., Grapevine, Winter triticale, Mixed cereal, Spring barley, Orchard, Sunflower, Sorghum, Durum wheat, Soybeans, Potatoes, Beet},
    xtick=data,
    nodes near coords align={vertical},
    axis y line*=left, 
    xticklabel style={rotate=50,anchor=east,font=\scriptsize},
    ymin=0,
    ymax=100,
    y label style={at={(axis description cs:+0.05,.5)},rotate=0,anchor=south},
    ]

\addplot [fill=gray] coordinates {(Meadow,94.4) (Winter wheat,63.7) (Corn,61.9) (Winter barley,46.4) (Leguminous,42.7) (Winter rapeseed,29.0) (Fruits\, veg.\, flow.,22.6) (Grapevine,20.0) (Winter triticale,19.8) (Mixed cereal,19.6) (Spring barley,19.4) (Orchard,17.7) (Sunflower,17.3) (Sorghum,15.3) (Durum wheat,15.1) (Soybeans,14.9) (Potatoes,12.7) (Beet,12.3)};
\legend{}; 
\end{axis}

\begin{axis}[
    width=.9\linewidth,
    height=.25\textheight,
    axis y line*=right, 
    axis x line=none,
    ymin=0,
    ymax=100,
    symbolic x coords={Meadow, Winter wheat, Corn, Winter barley, Leguminous, Winter rapeseed, Fruits\, veg.\, flow., Grapevine, Winter triticale, Mixed cereal, Spring barley, Orchard, Sunflower, Sorghum, Durum wheat, Soybeans, Potatoes, Beet},
    legend style={at={(0.5,-0.50)},
      anchor=north,legend columns=-1},
    ylabel={IoU}, 
    xticklabel style={rotate=90,anchor=east},
    ylabel near ticks, yticklabel pos=right,
]
\addlegendimage{/pgfplots/refstyle=Proportion}\addlegendentry{Proportion}
\addplot[mark=*,very thick, UTTCOLOR] coordinates {(Meadow,97.1) (Winter wheat,78.3) (Corn,77.5) (Winter barley,63.1) (Leguminous,60.0) (Winter rapeseed,57.5) (Fruits\, veg.\, flow.,50.6) (Grapevine,54.9) (Winter triticale,50.7) (Mixed cereal,36.9) (Spring barley,25.4) (Orchard,46.3) (Sunflower,49.2) (Sorghum,13.8) (Durum wheat,50.5) (Soybeans,59.0) (Potatoes,22.0) (Beet,69.9)};
\addlegendentry{UT\&T}
\addplot[mark=square*,SCALECOLOR, very thick] coordinates {(Meadow,97.1) (Winter wheat,84.2) (Corn,83.3) (Winter barley,56.4) (Leguminous,48.1) (Winter rapeseed,53.6) (Fruits\, veg.\, flow.,41.5) (Grapevine,54.9) (Winter triticale,27.5) (Mixed cereal,12.5) (Spring barley,15.1) (Orchard,30.4) (Sunflower,24.1) (Sorghum,2.6) (Durum wheat,43.9) (Soybeans,34.2) (Potatoes,12.3) (Beet,39.1)};
\addlegendentry{ScaleMAE}
\addplot[mark=diamond*,CROMACOLOR, very thick] coordinates {(Meadow,97.1) (Winter wheat,87.1) (Corn,88.9) (Winter barley,64.2) (Leguminous,53.3) (Winter rapeseed,86.5) (Fruits\, veg.\, flow.,61.3) (Grapevine,74.2) (Winter triticale,51.2) (Mixed cereal,22.4) (Spring barley,41.1) (Orchard,45.9) (Sunflower,64.5) (Sorghum,9.1) (Durum wheat,74.3) (Soybeans,59.1) (Potatoes,26.1) (Beet,76.5)};
\addlegendentry{CROMA}
\addplot[mark=triangle*,OMNICOLOR, very thick] coordinates {(Meadow,97.2) (Winter wheat,90.1) (Corn,92.2) (Winter barley,82.6) (Leguminous,56.9) (Winter rapeseed,89.0) (Fruits\, veg.\, flow.,56.7) (Grapevine,82.7) (Winter triticale,48.5) (Mixed cereal,34.8) (Spring barley,82.6) (Orchard,48.4) (Sunflower,75.0) (Sorghum,31.7) (Durum wheat,81.3) (Soybeans,85.7) (Potatoes,52.6) (Beet,88.1)};
\addlegendentry{\bf OmniSAT}
\end{axis}

\end{tikzpicture}
    \caption{{\bf Class-Wise Performance.} We plot the performance of different models for each class, sorted by decreasing frequency. OmniSat improves the performance across the board, and for rare classes in particular.}
    \label{fig:classwise}
\end{figure}

\begin{table}[t]
    \centering
    
\begin{center}
    TreeSatAI-TS \\
    \resizebox{\linewidth}{!}{\footnotesize{
    \begin{tabular}{lc*{16}{c}}
        
        
         Method & \begin{tabular}{c}  Macro  \\  F1  \end{tabular}   & \rotatebox{90}{\!\!\!\!Abies \rotatebox{-90}{\!\!\!\!\faTree}} & \rotatebox{90}{\!\!\!\!Acer \rotatebox{-90}{\!\!\!\!\tiny\faLeaf}} & \rotatebox{90}{\!\!\!\!Alnus \rotatebox{-90}{\!\!\!\!\tiny\faLeaf}} & \rotatebox{90}{\!\!\!\!Betula \rotatebox{-90}{\!\!\!\!\tiny\faLeaf}} & \rotatebox{90}{\!\!\!\!Cleared} & \rotatebox{90}{\!\!\!\!Fagus \rotatebox{-90}{\!\!\!\!\tiny\faLeaf}} & \rotatebox{90}{\!\!\!\!Fraxinus \rotatebox{-90}{\!\!\!\!\tiny\faLeaf}} & \rotatebox{90}{\!\!\!\!Larix \rotatebox{-90}{\!\!\!\!\faTree}} & \rotatebox{90}{\!\!\!\!Picea \rotatebox{-90}{\!\!\!\!\faTree}} & \rotatebox{90}{\!\!\!\!Pinus \rotatebox{-90}{\!\!\!\!\faTree}} & \rotatebox{90}{\!\!\!\!Populus \rotatebox{-90}{\!\!\!\!\tiny\faLeaf}} & \rotatebox{90}{\!\!\!\!Prunus \rotatebox{-90}{\!\!\!\!\tiny\faLeaf}} & \rotatebox{90}{\!\!\!\!Pseudotsuga \rotatebox{-90}{\!\!\!\!\faTree}} & \rotatebox{90}{\!\!\!\!Quercus \rotatebox{-90}{\!\!\!\!\tiny\faLeaf}} & \rotatebox{90}{\!\!\!\!Tilia \rotatebox{-90}{\!\!\!\!\tiny\faLeaf}}\\\greyrule
         \multicolumn{2}{l}{Proportion in \%} & 2.5 & 6.5 & 6.5 & 7.1 & 11.2 & 22.6 & 5.7 & 9.7 & 21.8 & 23.0 & 1.8 & 0.8 & 8.9 & 22.8 & 0.5 \\
        \midrule
        UT\&T & 48.8 & 52.8 & 43.4 & 49.8 & 36.8 & 53.7 & 60.7 & 45.9 & 43.7 & 63.6 & 72.1 & 44.8 & 32.4 & 45.4 & 62.5 & 25.0\\
        Scale-MAE & 47.3 & 53.2 & 32.4 & 42.7 & 38.5 & 64.4 & 63.8 & 38.3 & 51.7 & 71.5 & 78.2 & 25.4 & 11.8 & 55.1 & 64.0 & 18.2\\\greyrule
        OmniSat & \bf 73.4 & \bf 74.1 & \bf 58.9 & \bf 62.5 & \bf 56.5 & \bf 74.9 & \bf 72.9 & \bf 60.0 & \bf 69.5 & \bf 77.6 & \bf 85.4 & \bf 64.6 & \bf 46.2 & \bf 68.5 & \bf 76.1 & \bf 43.4 \\
    \end{tabular}
    }
    }\vspace{5mm}
    \\~
      {FLAIR}\\
  \resizebox{\linewidth}{!}{\footnotesize{ 
    \begin{tabular}{lc*{14}{c}}
      
        Method & \begin{tabular}{c}  Macro  \\  F1  \end{tabular}  & \rotatebox{90}{\!\!\!\!building} & \rotatebox{90}{\!\!\!\!perv. surface} & \rotatebox{90}{\!\!\!\!imperv. surface} & \rotatebox{90}{\!\!\!\!bare soil} & \rotatebox{90}{\!\!\!\!water} & \rotatebox{90}{\!\!\!\!coniferous} & \rotatebox{90}{\!\!\!\!deciduous} & \rotatebox{90}{\!\!\!\!brushwood} & \rotatebox{90}{\!\!\!\!vineyard} & \rotatebox{90}{\!\!\!\!herbaceous} &  \rotatebox{90}{\!\!\!\!agri. land} & \rotatebox{90}{\!\!\!\!plowed land} & \rotatebox{90}{\!\!\!\!other}\\\greyrule
         \multicolumn{2}{l}{Proportion in \%} & 27.6 & 45.0 & 37.7 & 8.0 & 12.5 & 36.0 & 84.2 & 49.6 & 6.5 & 74.1 & 39.4 & 7.1 & 11.0\\
        \midrule
        UT\&T & 57.3 & 52.4 & 63.3 & 60.4 & 32.7 & 46.9 & 64.4 & 91.2 & 66.3 & 57.1 & 84.7 & \bf 71.7 & 27.1 & 26.2\\
        Scale-MAE & 70.0 & 90.1 & 72.0 & 87.1 & 47.1 & \bf 81.3 & 65.1 & \bf 95.2 & 72.3 & 53.7 & \bf 88.8 & 70.2 & \bf 39.7 & 45.7\\\greyrule
        OmniSat & \bf 75.8 & \bf 91.8 & \bf 73.2 & \bf 88.2 & \bf 54.4 & \bf 81.3 & \bf 70.6 & \bf 95.2 & \bf 73.4 & \bf 72.1 & 87.5 & 71.1 & 36.0 & \bf 48.5\\
    \end{tabular}}}
\vspace{5mm}~\\
     {PASTIS-HD}\\

 \resizebox{\linewidth}{!}{\scriptsize{   
    \begin{tabular}{lc*{18}{c}}
        Method & \begin{tabular}{c}  Macro  \\  F1  \end{tabular} & \rotatebox{90}{\!\!\!\!Meadow} & \rotatebox{90}{\!\!\!\!Soft winter wheat} & \rotatebox{90}{\!\!\!\!Corn} & \rotatebox{90}{\!\!\!\!Winter barley} & \rotatebox{90}{\!\!\!\!Winter rapeseed} & \rotatebox{90}{\!\!\!\!Spring barley} & \rotatebox{90}{\!\!\!\!Sunflower} & \rotatebox{90}{\!\!\!\!Grapevine} & \rotatebox{90}{\!\!\!\!Beet} &  \rotatebox{90}{\!\!\!\!Winter triticale} & \rotatebox{90}{\!\!\!\!Winter durum wheat} & \rotatebox{90}{\!\!\!\!Fruits, vegetables, flowers} & \rotatebox{90}{\!\!\!\!Potatoes} & \rotatebox{90}{\!\!\!\!Leguminous fodder} & \rotatebox{90}{\!\!\!\!Soybeans}& \rotatebox{90}{\!\!\!\!Orchard} & \rotatebox{90}{\!\!\!\!Mixed cereal}& \rotatebox{90}{\!\!\!\!Sorghum}\\\greyrule
         \multicolumn{2}{l}{Proportion in \%} & 94.4 & 63.7 & 61.9 & 46.4 & 29.0 & 19.4 & 17.3 & 20.0 & 12.3 & 19.8 & 15.1 & 22.6 & 12.7 & 42.7 & 14.9 & 17.7 & 19.6 & 15.3\\
        \midrule
        UT\&T & 53.5 & 97.1 & 78.3 & 77.5 & 63.1 & 57.5 & 25.4 & 49.2 & 54.9 & 69.9 & 50.7 & 50.5 & 50.6 & 22.0 & \bf 60.0 & 59.0 & 46.3 & \bf 36.9 & 13.8 \\
        CROMA & 60.1 & 97.1 & 87.1 & 88.9 & 64.2 & 86.5 & 41.1 & 64.5 & 74.2 & 76.5 & \bf 51.2 & 74.3 & \bf 61.3 & 26.1 & 53.3 & 59.1 & 45.9 & 22.4 & 9.1\\
        Scale-MAE & 42.2 & 97.1 & 84.2 & 83.3 & 56.4 & 53.6 & 15.1 & 24.1 & 54.9 & 39.1 & 27.5 & 43.9 & 41.5 & 12.3 & 48.1 & 34.2 & 30.4 & 12.5 & 2.6\\\greyrule
        OmniSat & \bf 69.9 & \bf 97.2 & \bf 90.1 & \bf 92.2 & \bf 82.6 & \bf 89.0 & \bf 64.7 & \bf 75.0 & \bf 82.7 & \bf 88.1 & 48.5 & \bf 81.3 & 56.7 & \bf 52.6 & 56.9 & \bf 85.7 & \bf 48.4 & 34.8 & \bf 31.7 \\
    \end{tabular}}}
\end{center}

    \caption{{\bf Class-Wise Performance.} We report the F1-score for each class for TreeSatAI-TS, FLAIR, and PASTIS-HD  for multilabel classification. We also report the unweighted class-averaged F1-score (Macro-F1).  We can observe that OmniSat outperforms UT\&T \cite{garioud2023flair} and Scale-MAE \cite{reed2023scale} on nearly all classes for both datasets. In particular, we observe the most significant gains for classes with discriminative temporal dynamics, such as broadleaf tree species and the vineyards class.}
    \label{tab:classwise}
\end{table}

\paragraph{\bf Failure Case.}
We report in the bottom half of \figref{fig:quali} hard examples from our three datasets and compare the prediction of OmniSat and other models.
For the TreeSatAI-TS example, the Sentinel-2 optical time-series is highly occluded: over 80\% of acquisitions are covered by clouds. Furthermore, the forest tile contains a large variety of tree species organized in densely connected canopy, making its classification particularly hard. Indeed, the texture of the images in closed forests does not bring additional discriminative information. 

The example from FLAIR is a scrap yard, which is almost entirely covered by broken vehicles. Since FLAIR's annotations focus on the ground rather than transient or stationary objects, identifying the actual land cover in such scenarios is very challenging. 

The image taken from PASTIS contains a mix of several different crop types, including the class \emph{mixed cereal} which can already correspond to a parcel with various cereal types. This leads to a hard classification problem for all methods.

\begin{figure}[t]
    \centering
    \begin{tabular}{l@{\;\;}l@{\;\;\;\;}l@{\;\;}l@{\;\;}l@{\;\;}l@{\;\;}l}
    &\multicolumn{2}{c}{\scriptsize  Inputs} & \scriptsize Ground truth & \bf \scriptsize \underline{OmniSat} &  \scriptsize UT\&T \cite{garioud2023flair} & \scriptsize Scale-\scriptsize MAE \cite{reed2023scale} \\
    \rotatebox{90}{\;\;\;\;\scriptsize  TreeSatAI-TS}
     &
      \includegraphics[width=.17\linewidth, angle=180,origin=c]{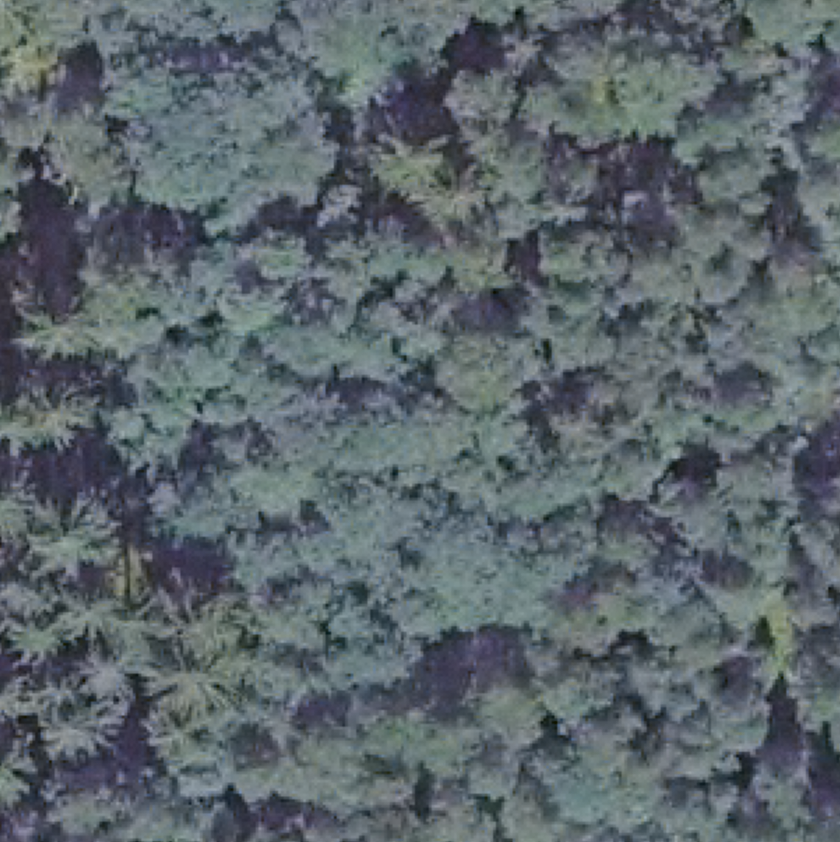}
    &
     \begin{minipage}[t][.01\linewidth][t]{.13\linewidth}
     \vspace{-20.5mm}\centering
    \begin{tabular}{c@{\;}c}
        \includegraphics[width=.38\linewidth]{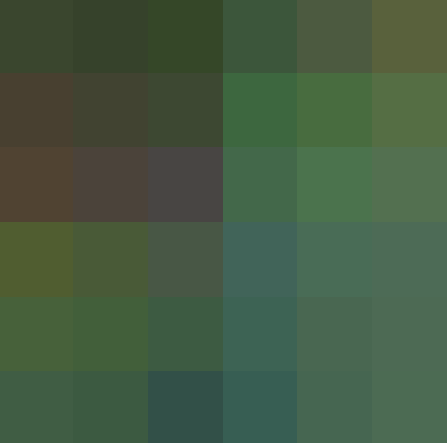}
        &
        \includegraphics[width=.38\linewidth]{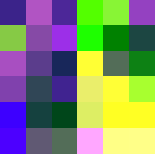}
       \\
       \includegraphics[width=.38\linewidth]{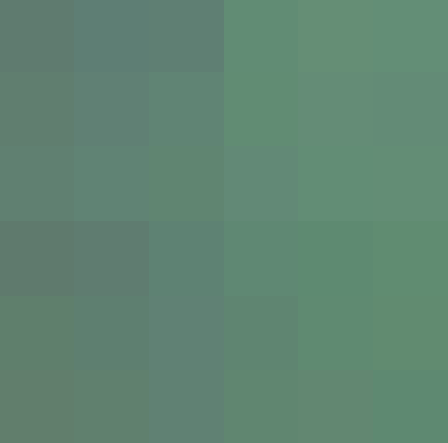}
       &
    \includegraphics[width=.38\linewidth]{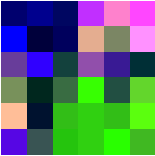}
       \\
       \includegraphics[width=.38\linewidth]{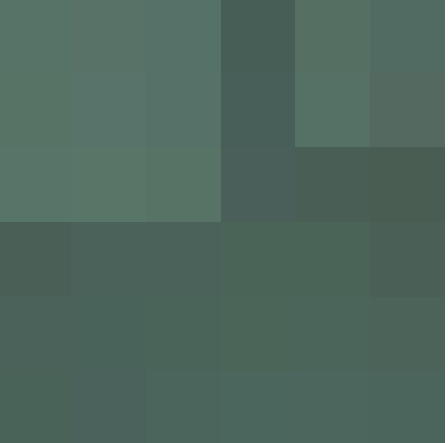}
       &
    \includegraphics[width=.38\linewidth]{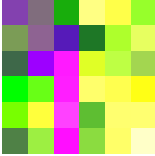}
    \end{tabular}
    \end{minipage}
     &
     \begin{minipage}[t][.01\linewidth][t]{.15\linewidth}
     \vspace{-18mm}\scriptsize {
     \begin{tabular}{m{\linewidth}}
        - Picea \faTree\\
        - Betula \tiny \faLeaf\\
        - Alnus \tiny \faLeaf\\
         - Quercus \tiny \faLeaf
     \end{tabular}}
      \end{minipage}
          &
     \begin{minipage}[t][.01\linewidth][t]{.15\linewidth}
     \vspace{-18mm}\scriptsize {
     \begin{tabular}{m{\linewidth}}
        - \textcolor{green!50!black}{Picea} \\
        - \textcolor{green!50!black}{Betula} \\
        - \textcolor{green!50!black}{Alnus}  \\
        - \textcolor{red!50!black}{\xmark}
     \end{tabular}}
      \end{minipage}
      &
     \begin{minipage}[t][.01\linewidth][t]{.15\linewidth}
     \vspace{-18mm}\scriptsize {
     \begin{tabular}{m{\linewidth}}
         - \textcolor{green!50!black}{Picea} \\
         - \textcolor{green!50!black}{Betula} \\
         - \textcolor{green!50!black}{Alnus}  \\
         - \textcolor{red!50!black}{\xmark}\\
         - \textcolor{red!50!black}{Pinus} \faTree\\
     \end{tabular}}
      \end{minipage}
     &
     \begin{minipage}[t][.01\linewidth][t]{.15\linewidth}
     \vspace{-18mm}\scriptsize {
     \begin{tabular}{m{\linewidth}}
         - \textcolor{green!50!black}{Picea} \\
         - \textcolor{red!50!black}{\xmark} \\
         - \textcolor{red!50!black}{\xmark} \\
         - \textcolor{red!50!black}{\xmark}
     \end{tabular}}
      \end{minipage}
     \\~\\
    \rotatebox{90}{\;\qquad\scriptsize  FLAIR}
     &
      \includegraphics[width=.17\linewidth, angle=0]{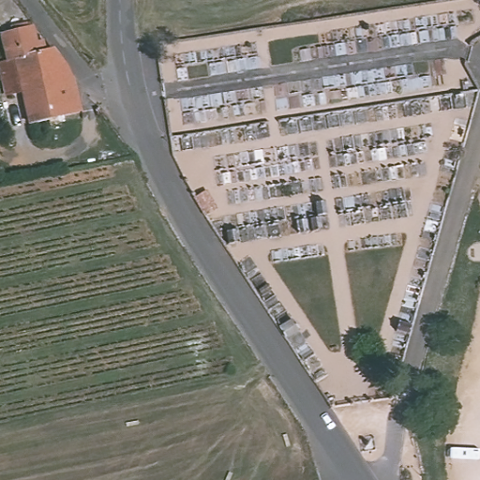}
    &
     \begin{minipage}[t][.01\linewidth][t]{.13\linewidth}
     \vspace{-20.5mm}\centering
    \begin{tabular}{@{\;}c}
        \includegraphics[width=.38\linewidth,angle=180, origin=c]{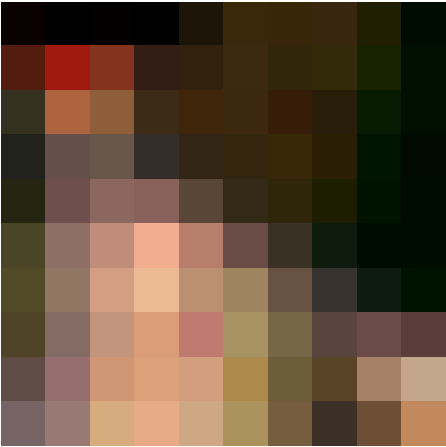}
       \\
       \includegraphics[width=.38\linewidth,angle=180, origin=c]{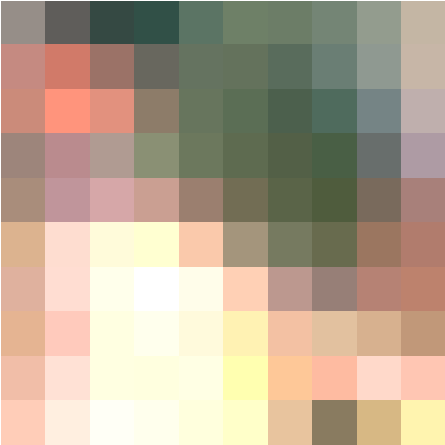}
       \\
    \includegraphics[width=.38\linewidth,angle=180, origin=c]{images/flair_s21.png}
    \end{tabular}
    \end{minipage}
     &
     \begin{minipage}[t][.01\linewidth][t]{.15\linewidth}
     \vspace{-20.5mm}\scriptsize {
     \begin{tabular}{m{\linewidth}}
         - building  \\
         - pervious surf.\\
         - impervious surf.\\
         - deciduous\\
         - brushwood\\
         - herbaceous\\
         - agricultural \\
         - vineyard
     \end{tabular}}
      \end{minipage}
          &
     \begin{minipage}[t][.01\linewidth][t]{.15\linewidth}
     \vspace{-20.5mm}\scriptsize {
     \begin{tabular}{m{\linewidth}}
         - \textcolor{green!50!black}{building}  \\
         - \textcolor{green!50!black}{pervious surf.}\\
         - \textcolor{green!50!black}{impervious surf.}\\
         - \textcolor{green!50!black}{deciduous}\\
         - \textcolor{green!50!black}{brushwood}\\
         - \textcolor{green!50!black}{herbaceous}\\
         - \textcolor{green!50!black}{agricultural} \\
         - \textcolor{green!50!black}{vineyard}
     \end{tabular}}
      \end{minipage}
         &
     \begin{minipage}[t][.01\linewidth][t]{.15\linewidth}
     \vspace{-20.5mm}\scriptsize {
     \begin{tabular}{m{\linewidth}}
         - \textcolor{green!50!black}{building}  \\
         - \textcolor{green!50!black}{pervious surf.}\\
         - \textcolor{green!50!black}{impervious surf.}\\
         - \textcolor{green!50!black}{deciduous}\\
         - \textcolor{green!50!black}{brushwood}\\
         - \textcolor{green!50!black}{herbaceous}\\
         - \textcolor{green!50!black}{agricultural} \\
         - \textcolor{red!50!black}{\xmark}\\
     \end{tabular}}
      \end{minipage} 
          &
     \begin{minipage}[t][.01\linewidth][t]{.15\linewidth}
     \vspace{-20.5mm}\scriptsize {
     \begin{tabular}{m{\linewidth}}
         - \textcolor{green!50!black}{building}  \\
         - \textcolor{green!50!black}{pervious surf.}\\
         - \textcolor{green!50!black}{impervious surf.}\\
         - \textcolor{green!50!black}{deciduous}\\
         - \textcolor{green!50!black}{brushwood}\\
         - \textcolor{green!50!black}{herbaceous}\\
         - \textcolor{red!50!black}{\xmark}\\
         - \textcolor{red!50!black}{\xmark}
     \end{tabular}}
      \end{minipage}
      \\~\\
    \rotatebox{90}{\;\qquad\scriptsize PASTIS-HD}
     &
      \includegraphics[width=.17\linewidth, angle=0]{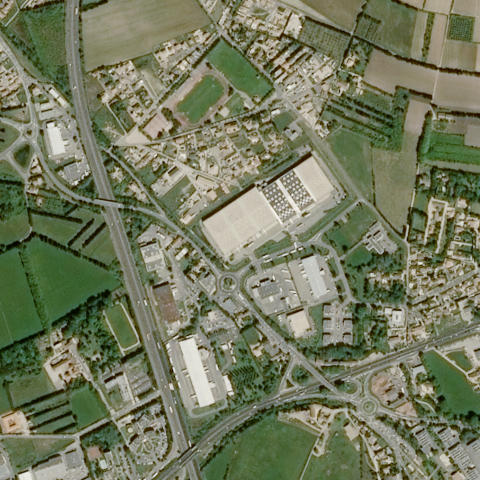}
    &
     \begin{minipage}[t][.01\linewidth][t]{.13\linewidth}
     \vspace{-20.5mm}\centering
    \begin{tabular}{c@{\;}c}
        \includegraphics[width=.38\linewidth]{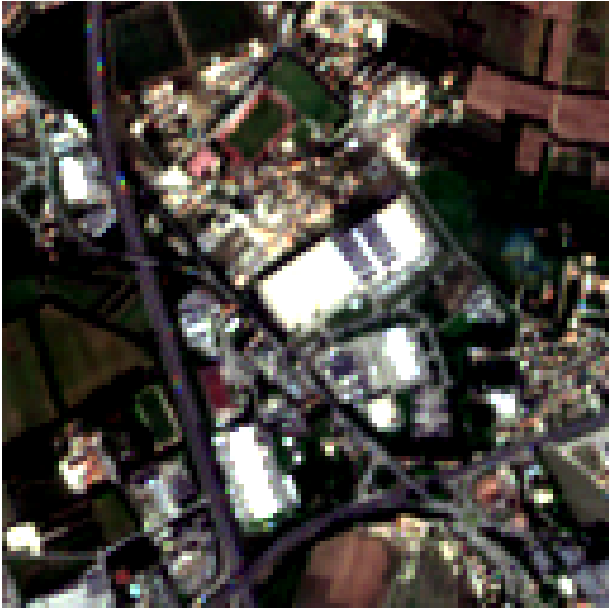}
        &
        \includegraphics[width=.38\linewidth]{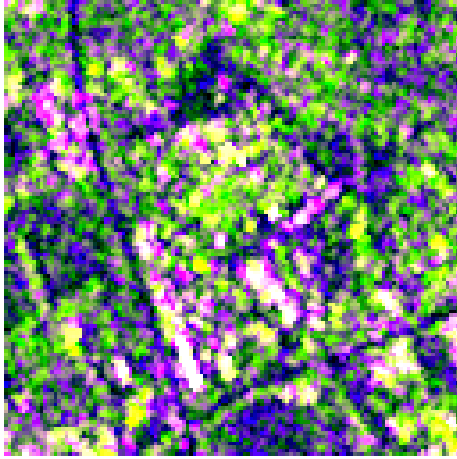}
       \\
       \includegraphics[width=.38\linewidth]{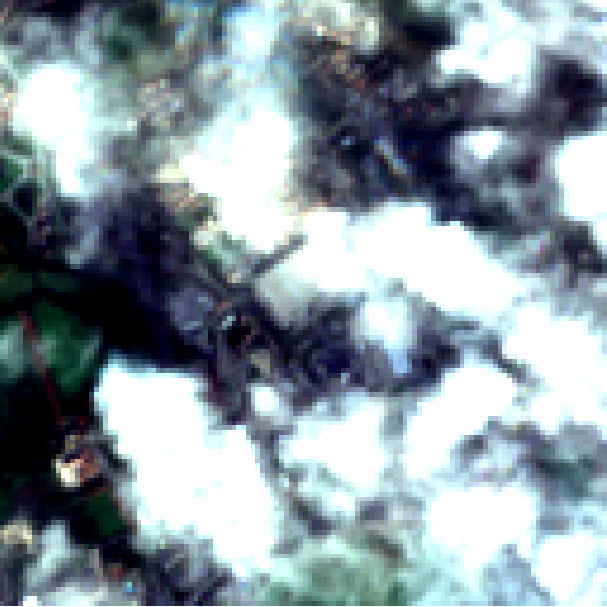}
       &
    \includegraphics[width=.38\linewidth]{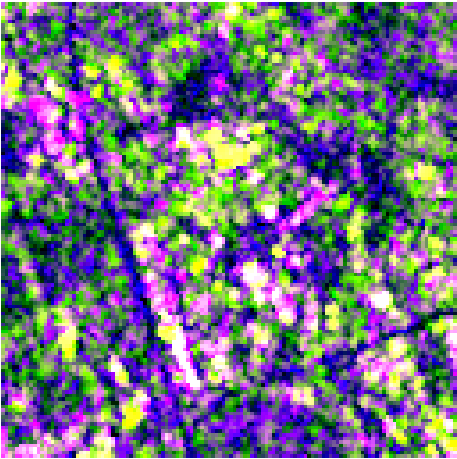}
       \\
       \includegraphics[width=.38\linewidth]{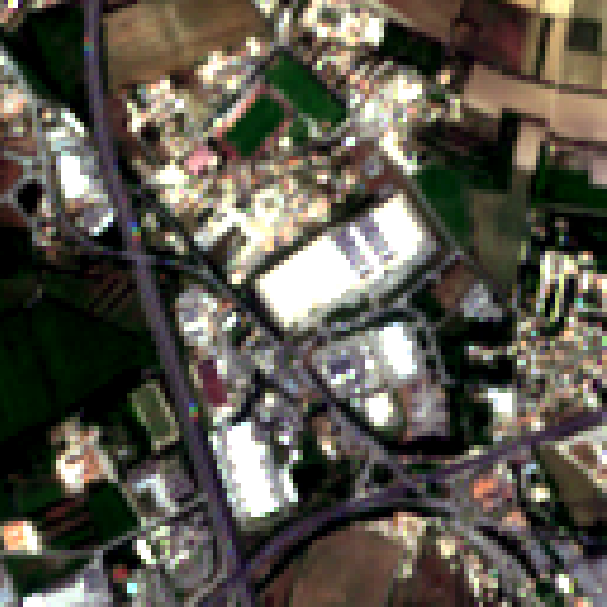}
       &
    \includegraphics[width=.38\linewidth]{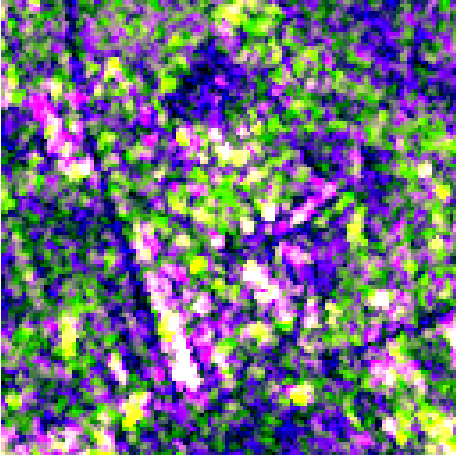}
    \end{tabular}
    \end{minipage}
     &
     \begin{minipage}[t][.01\linewidth][t]{.15\linewidth}
     \vspace{-17.5mm}\scriptsize {
     \begin{tabular}{m{\linewidth}}
         - Meadow \\
         - Soft winter wheat \\
         - Corn \\
         - Winter rapeseed \\
         - Beet
     \end{tabular}}
      \end{minipage}
          &
     \begin{minipage}[t][.01\linewidth][t]{.15\linewidth}
     \vspace{-17.5mm}\scriptsize {
     \begin{tabular}{m{\linewidth}}
         - \textcolor{green!50!black}{Meadow} \\
         - \textcolor{green!50!black}{Soft winter wheat} \\
         - \textcolor{green!50!black}{Corn} \\
         - \textcolor{green!50!black}{Winter rapeseed} \\
         - \textcolor{green!50!black}{Beet}
     \end{tabular}}
      \end{minipage}
         &
     \begin{minipage}[t][.01\linewidth][t]{.15\linewidth}
     \vspace{-17.5mm}\scriptsize {
     \begin{tabular}{m{\linewidth}}
         - \textcolor{green!50!black}{Meadow} \\
         - \textcolor{red!50!black}{\xmark}\\
         - \textcolor{red!50!black}{\xmark}\\
         - \textcolor{red!50!black}{\xmark}\\
         - \textcolor{red!50!black}{\xmark}\\
         - \textcolor{red!50!black}{Potatoes}
     \end{tabular}}
      \end{minipage} 
          &
     \begin{minipage}[t][.01\linewidth][t]{.15\linewidth}
     \vspace{-17.5mm}\scriptsize {
     \begin{tabular}{m{\linewidth}}
         - \textcolor{green!50!black}{Meadow} \\
         - \textcolor{red!50!black}{\xmark}\\
         - \textcolor{red!50!black}{\xmark}\\
         - \textcolor{red!50!black}{\xmark}\\
         - \textcolor{red!50!black}{\xmark}\\
         - \textcolor{red!50!black}{Sunflower}\\
         - \textcolor{red!50!black}{Grapevine}
     \end{tabular}}
      \end{minipage} \\
\midrule
    \rotatebox{90}{\;\;\;\;\scriptsize  TreeSatAI-TS}
     &
      \includegraphics[width=.17\linewidth, angle=180,origin=c]{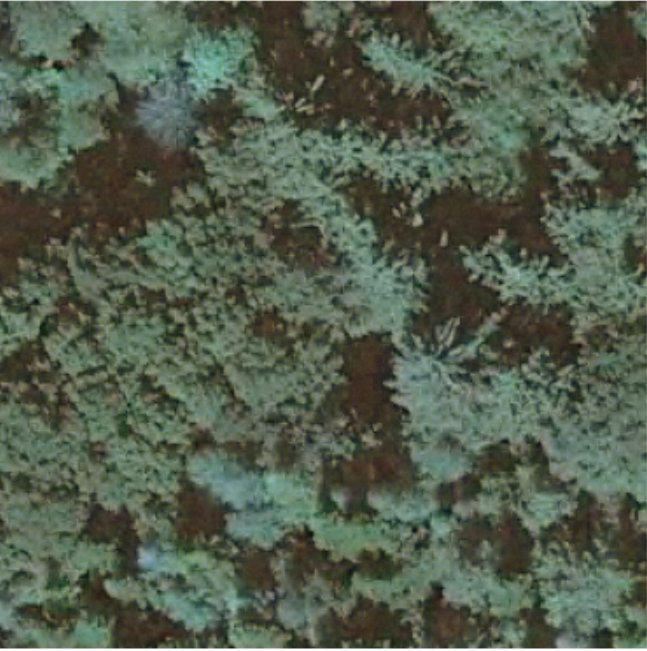}
    &
     \begin{minipage}[t][.01\linewidth][t]{.13\linewidth}
     \vspace{-20.5mm}\centering
    \begin{tabular}{c@{\;}c}
        \includegraphics[width=.38\linewidth]{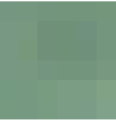}
        &
        \includegraphics[width=.38\linewidth]{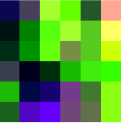}
       \\
       \includegraphics[width=.38\linewidth]{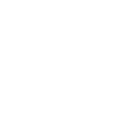}
       &
    \includegraphics[width=.38\linewidth]{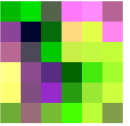}
       \\
       \includegraphics[width=.38\linewidth]{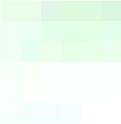}
       &
    \includegraphics[width=.38\linewidth]{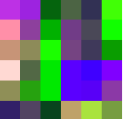}
    \end{tabular}
    \end{minipage}
     &
     \begin{minipage}[t][.01\linewidth][t]{.13\linewidth}
     \vspace{-21mm}\scriptsize {
     \begin{tabular}{m{\linewidth}}
        - Quercus \tiny \faLeaf\\
        - Acer \tiny \faLeaf\\
        - Alnus \tiny \faLeaf\\
        - Larix \tiny \faTree\\
     \end{tabular}}
      \end{minipage}
          &
     \begin{minipage}[t][.01\linewidth][t]{.13\linewidth}
     \vspace{-21mm}\scriptsize {
     \begin{tabular}{m{\linewidth}}
        - \textcolor{green!50!black}{Quercus}  \\
        - \textcolor{red!50!black}{\xmark} \\
        - \textcolor{red!50!black}{\xmark}\\
        - \textcolor{red!50!black}{\xmark} \\
        - \textcolor{red!50!black}{Abies} \tiny \faTree\\
     \end{tabular}}
      \end{minipage}
      &
     \begin{minipage}[t][.01\linewidth][t]{.13\linewidth}
     \vspace{-21mm}\scriptsize {
     \begin{tabular}{m{\linewidth}}
         - \textcolor{red!50!black}{\xmark}\\
         - \textcolor{red!50!black}{\xmark}\\
         - \textcolor{red!50!black}{\xmark}\\
         - \textcolor{red!50!black}{\xmark} \\
         - \textcolor{red!50!black}{Abies} \tiny \faTree \\
         - \textcolor{red!50!black}{Betula}  \tiny \faLeaf
     \end{tabular}}
      \end{minipage}
     &
     \begin{minipage}[t][.01\linewidth][t]{.13\linewidth}
     \vspace{-21mm}\scriptsize {
     \begin{tabular}{m{\linewidth}}
         - \textcolor{red!50!black}{\xmark} \\
         - \textcolor{red!50!black}{\xmark} \\
         - \textcolor{red!50!black}{\xmark} \\
         - \textcolor{red!50!black}{\xmark} \\
         - \textcolor{red!50!black}{Picea} \tiny \faTree\\
     \end{tabular}}
      \end{minipage}
     \\~\\
    \rotatebox{90}{\;\qquad\scriptsize  FLAIR}
     &
      \includegraphics[width=.17\linewidth, angle=0]{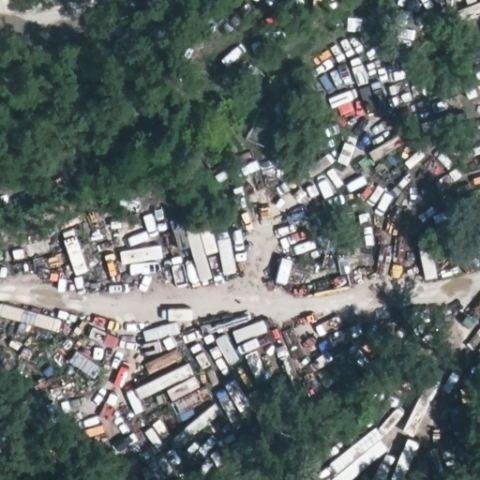}
    &
     \begin{minipage}[t][.01\linewidth][t]{.13\linewidth}
     \vspace{-20.5mm}\centering
    \begin{tabular}{@{\;}c}
        \includegraphics[width=.38\linewidth,angle=0, origin=c]{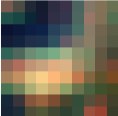}
       \\
       \includegraphics[width=.38\linewidth,angle=0, origin=c]{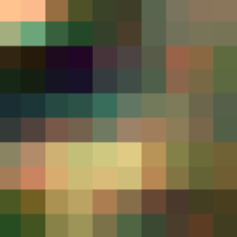}
       \\
    \includegraphics[width=.38\linewidth,angle=0, origin=c]{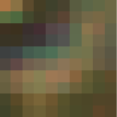}
    \end{tabular}
    \end{minipage}
     &
     \begin{minipage}[t][.01\linewidth][t]{.13\linewidth}
     \vspace{-24.5mm}\scriptsize {
     \begin{tabular}{m{\linewidth}}
         - decideous \\
         - herbaceous\\
         - water \\
         - pervious surf.\\
         - bare soil
     \end{tabular}}
      \end{minipage}
          &
     \begin{minipage}[t][.01\linewidth][t]{.13\linewidth}
     \vspace{-24.5mm}\scriptsize {
     \begin{tabular}{m{\linewidth}}
         - \textcolor{green!50!black}{decideous}\\
         - \textcolor{green!50!black}{herbaceous}\\
         - \textcolor{green!50!black}{water}\\
         - \textcolor{red!50!black}{\xmark}\\
         - \textcolor{red!50!black}{\xmark}\\
         - \textcolor{red!50!black}{building}\\
         - \textcolor{red!50!black}{imperv. surf.}\\
         - \textcolor{red!50!black}{brushwood}
     \end{tabular}}
      \end{minipage}
         &
     \begin{minipage}[t][.01\linewidth][t]{.13\linewidth}
     \vspace{-24.5mm}\scriptsize {
     \begin{tabular}{m{\linewidth}}
         - \textcolor{green!50!black}{decideous}\\
         - \textcolor{green!50!black}{herbaceous}\\
         - \textcolor{red!50!black}{\xmark}\\
         - \textcolor{green!50!black}{pervious surf.}\\
         - \textcolor{red!50!black}{\xmark}\\
         - \textcolor{red!50!black}{building}\\
         - \textcolor{red!50!black}{imperv. surf.}\\
         - \textcolor{red!50!black}{brushwood}\\
         - \textcolor{red!50!black}{coniferous}
     \end{tabular}}
      \end{minipage} 
          &
     \begin{minipage}[t][.01\linewidth][t]{.13\linewidth}
     \vspace{-24.5mm}\scriptsize {
     \begin{tabular}{m{\linewidth}}
         - \textcolor{green!50!black}{decideous}\\
         - \textcolor{green!50!black}{herbaceous}\\
         - \textcolor{red!50!black}{\xmark}\\
         - \textcolor{green!50!black}{pervious surf.}\\
         - \textcolor{red!50!black}{\xmark}\\
         - \textcolor{red!50!black}{building}\\
         - \textcolor{red!50!black}{imperv. surf.}\\
         - \textcolor{red!50!black}{brushwood}\\
         - \textcolor{red!50!black}{coniferous}\\
         - \textcolor{red!50!black}{other}
     \end{tabular}}
      \end{minipage}
     \\~\\
    \rotatebox{90}{\;\;\;\;\;\;\scriptsize PASTIS-HD}
     &
      \includegraphics[width=.17\linewidth, angle=0]{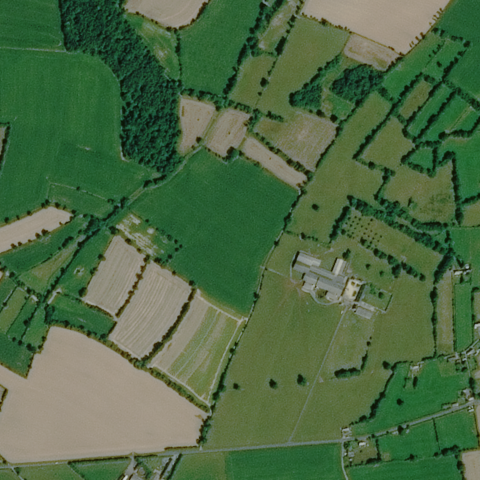}
    &
     \begin{minipage}[t][.01\linewidth][t]{.13\linewidth}
     \vspace{-20.5mm}\centering
    \begin{tabular}{c@{\;}c}
        \includegraphics[width=.38\linewidth]{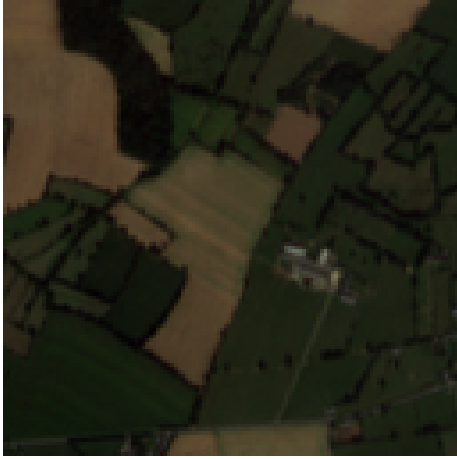}
        &
        \includegraphics[width=.38\linewidth]{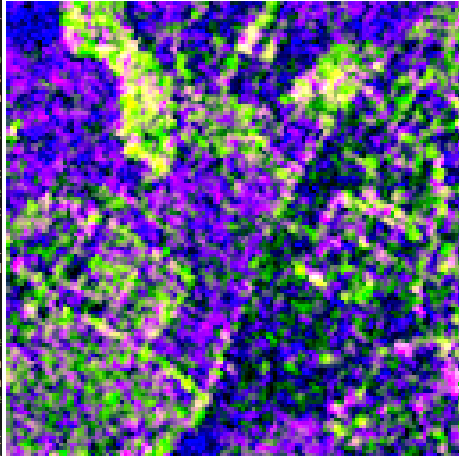}
       \\
       \includegraphics[width=.38\linewidth]{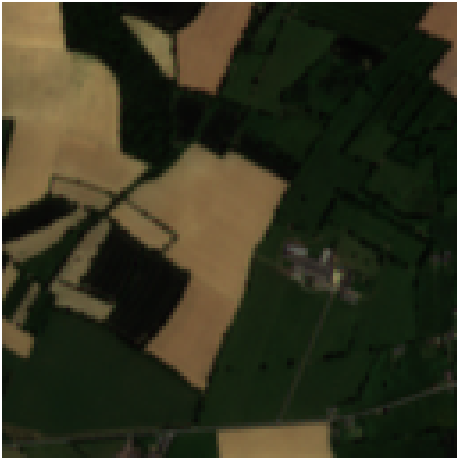}
       &
    \includegraphics[width=.38\linewidth]{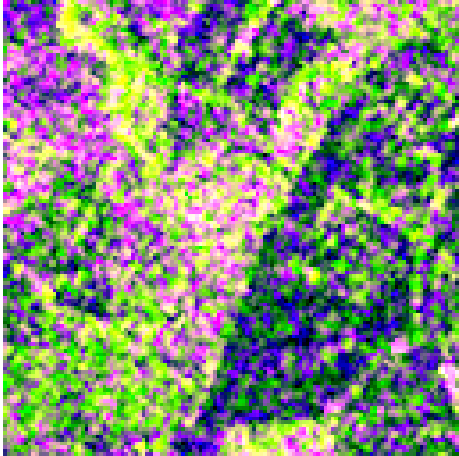}
       \\
       \includegraphics[width=.38\linewidth]{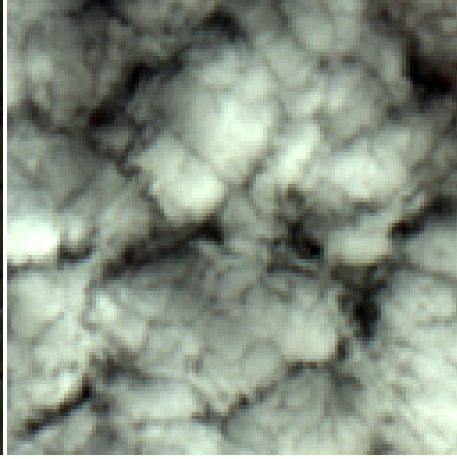}
       &
    \includegraphics[width=.38\linewidth]{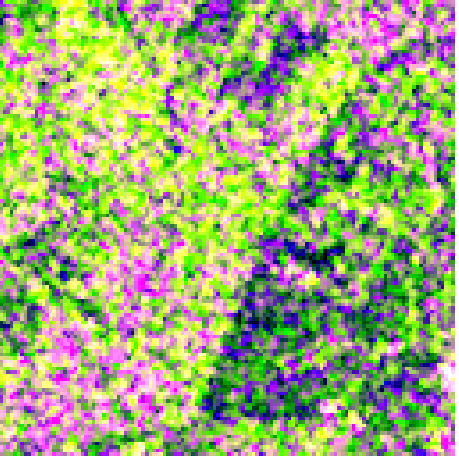}
    \end{tabular}
    \end{minipage}
     &
     \begin{minipage}[t][.01\linewidth][t]{.13\linewidth}
     \vspace{-21.4mm}\scriptsize {
     \begin{tabular}{m{\linewidth}}
         - Meadow \\
         - Winter wheat \\
         - Corn \\
         - Potatoes \\
         - Mixed cereal
     \end{tabular}}
      \end{minipage}
          &
     \begin{minipage}[t][.01\linewidth][t]{.13\linewidth}
     \vspace{-21.4mm}\scriptsize {
     \begin{tabular}{m{\linewidth}}
         - \textcolor{green!50!black}{Meadow} \\
         - \textcolor{green!50!black}{Winter wheat} \\
         - \textcolor{green!50!black}{Corn} \\
         - \textcolor{green!50!black}{Potatoes} \\
         - \textcolor{red!50!black}{\xmark} \\
         - \textcolor{red!50!black}{Winter barley}\\
         - \textcolor{red!50!black}{Wint. rapeseed}\\
         - \textcolor{red!50!black}{Legum. fodder}\\
     \end{tabular}}
      \end{minipage}
         &
     \begin{minipage}[t][.01\linewidth][t]{.13\linewidth}
     \vspace{-21.4mm}\scriptsize {
     \begin{tabular}{m{\linewidth}}
         - \textcolor{green!50!black}{Meadow} \\
         - \textcolor{red!50!black}{\xmark}\\
         - \textcolor{red!50!black}{\xmark}\\
         - \textcolor{red!50!black}{\xmark}\\
         - \textcolor{red!50!black}{\xmark}\\
         - \textcolor{red!50!black}{Spring barley}\\
         - \textcolor{red!50!black}{Orchard}\\
         - \textcolor{red!50!black}{Legum. fodder}\\
         - \textcolor{red!50!black}{Durum wheat}\\
         - \textcolor{red!50!black}{Fruits, veg..}
     \end{tabular}}
      \end{minipage} 
          &
     \begin{minipage}[t][.01\linewidth][t]{.13\linewidth}
     \vspace{-21.4mm}\scriptsize {
     \begin{tabular}{m{\linewidth}}
         - \textcolor{green!50!black}{Meadow} \\
         - \textcolor{green!50!black}{Winter wheat} \\
         - \textcolor{green!50!black}{Corn} \\
         - \textcolor{red!50!black}{\xmark} \\
         - \textcolor{red!50!black}{\xmark} \\
         - \textcolor{red!50!black}{Winter barley}\\
         - \textcolor{red!50!black}{Wint. rapeseed}\\
         - \textcolor{red!50!black}{Winter triticale}\\
     \end{tabular}}
      \end{minipage}
      ~\\~\\
\end{tabular}

    \caption{{\bf Qualitative Results.} We report predictions of OmniSat and two competing models on tiles from our datasets, including a failure case (bottom). OmniSat can detect classes with recognizable temporal dynamics such as agricultural lands or mixed forest areas with both coniferous \scriptsize{\faTree}\small~and deciduous trees \tiny \raisebox{0.2mm}{\faLeaf}\small. Other methods, and in particular ScaleMAE, struggle to detect these classes.}
    \label{fig:quali}
\end{figure}

 }{}

\end{document}